\documentclass{article}

\usepackage{microtype}
\usepackage{graphicx}
\usepackage{subfigure}
\usepackage{booktabs} 

\usepackage{hyperref}

\usepackage[accepted]{icml2025}

\usepackage{amsmath}
\usepackage{amssymb}
\usepackage{mathtools}
\usepackage{amsthm}

\usepackage[capitalize,noabbrev]{cleveref}

\usepackage[textsize=tiny]{todonotes}

\icmltitlerunning{Value from Observations}

\newcommand{\background}{\emph{background}}
\newcommand{\expert}{\emph{expert}}

\begin{document}

\twocolumn[
\icmltitle{Value from Observations: \\ Towards Large-Scale Imitation Learning via Self-Improvement}



\icmlsetsymbol{equal}{*}

\begin{icmlauthorlist}
\icmlauthor{Michael Bloesch}{gdm}
\icmlauthor{Markus Wulfmeier}{gdm}
\icmlauthor{Philemon Brakel}{gdm}
\icmlauthor{Todor Davchev}{gdm}
\icmlauthor{Martina Zambelli}{gdm}
\icmlauthor{Jost Tobias Springenberg}{gdm}
\icmlauthor{Abbas Abdolmaleki}{gdm}
\icmlauthor{William F Whitney}{gdm}
\icmlauthor{Nicolas Heess}{gdm}
\icmlauthor{Roland Hafner}{gdm}
\icmlauthor{Martin Riedmiller}{gdm}
\end{icmlauthorlist}

\icmlaffiliation{gdm}{Google Deepmind, London, United Kingdom}
\icmlcorrespondingauthor{Michael Bloesch}{bloesch@google.com}

\icmlkeywords{Imitation learning from observation, self-improvement}

\vskip 0.3in
]



\printAffiliationsAndNotice{}  

\begin{abstract}

Imitation Learning from Observation (IfO) offers a powerful way to learn behaviors at large-scale: Unlike behavior cloning or offline reinforcement learning, IfO can leverage action-free demonstrations and thus circumvents the need for costly action-labeled demonstrations or reward functions.
However, current IfO research focuses on idealized scenarios with mostly bimodal-quality data distributions, restricting the meaningfulness of the results.
In contrast, this paper investigates more nuanced distributions and introduces a method to learn from such data, moving closer to a paradigm in which imitation learning can be performed iteratively via self-improvement.
Our method adapts RL-based imitation learning to action-free demonstrations, using a value function to transfer information between expert and non-expert data.
Through comprehensive evaluation, we delineate the relation between different data distributions and the applicability of algorithms and highlight the limitations of established methods.
Our findings provide valuable insights for developing more robust and practical IfO techniques on a path to scalable behaviour learning.

\end{abstract}

\section{Introduction}

Bringing the cognitive capabilities of large vision and language models to embodied systems is at the forefront of many researchers' attention. Given the nature of the underlying data, robotics has been a popular target domain with a variety of deployed approaches, including prompting based algorithms \citep{ahn2022icanisay,jiang2023vima,kwon2024language,di2023towards} and behavior cloning (BC) \citep{Brohan2022RT1RT,bousmalis2024robocat}. We are motivated by a practical future where agents are trained on a large-scale dataset of language and video without requiring explicit action annotations. The agent should autonomously collect data to understand its action space and address knowledge gaps as required. Enabling learning in such a setting overcomes two key hurdles to scaling imitation learning: the static limitations of prompt engineering that may hinder generalization and the high cost of large-scale action-labeled demonstrations. 

As a first step towards this setting, this paper studies the problem of training an agent assuming access to two kinds of data: \expert{} demonstrations without action labels, and a possibly self-collected \background{} dataset which includes actions but does not necessarily match expert behaviour.
So while the agent has access to some action data, it has not directly experienced a task being solved (it may only see solutions within the action-free \expert{} data) and does not observe any external rewards. 
On a smaller scale, this setup is found when expert action labels are difficult to obtain or in cross-embodiment scenarios, e.g. in autonomous driving \citep{dosovitskiy2017carla} or the UMI-gripper \citep{chi2024universal}.
Expanding the \expert{} data to include in-the-wild observations of humans solving tasks would allow scaling to cheaply available large-scale data.

Any applicable method has to address two key challenges: a) collect useful data for imitating the expert behaviour and b) learn a robust imitation policy from that data.
Collecting data and inferring policies is at the heart of reinforcement learning (RL). Driven by a reward function, RL agents collect data and improve a policy based on the collected, possibly non-expert, experience. The reward serves as key signal for improvement and represents the main judge of \emph{good} and \emph{bad}. However, engineering a reward model can be time-intensive \citep{Tirumala2024LearningRS, openai2019dota, anymalwalk} and cannot capture the breadth of possible contexts and tasks encountered at scale. We are thus interested in algorithms that acquire a notion of \emph{good} and \emph{bad} directly from data.
This brings us to imitation learning, in particular Imitation learning from Observations (IfO) which addresses the problem of imitating expert behavior without requiring action annotations \citep{torabi2018behavioral,torabi2018generative,liu2018imitation}. 

To date, IfO remains an open research problem \citep{sikchi2024dual} and current methods do not match the maturity of related methods in RL, which have been shown to scale to large datasets and models \citep{ chebotar2021actionable,springenberg2024offline}. In contrast, IfO methods have been benchmarked only in small, single-domain settings with ad-hoc choices for data generation \citep{pmlr-v162-ma22a,zolna2020offline,sikchi2024dual}. We here suggest a concrete benchmark that moves towards more nuanced \background{} data compositions that better reflect the above large-scale vision.
Previous work has mainly focused on \background{} data where a significant amount of expert data was diluted in non-expert data \citep{pmlr-v162-ma22a,zolna2020offline,sikchi2024dual}. Instead, we extend previous datasets \citep{fu2021d4rl,robomimic2021} and collect data with a variety of policies of different quality and examine how much performance can be improved across this spectrum. We argue that this new self-improvement benchmark (SIBench) more accurately reflects the scenario where an agent collects its own data for improvement and we compare this with an iterative self-improvement experiment where an agent uses self-collected data to improve.

In addition to forming a new benchmark, we propose a simple, offline IfO method.
Our algorithm adapts either SQIL~\citep{reddy2020sqil} or ORIL~\citep{zolna2020offline}, two RL-based imitation learning algorithms, to the action free setting. For the SQIL based variant, we assign a reward of 1 to \expert{} data and a reward of 0 to \background{} data before applying a value-function based offline RL method akin to AWR \citep{wang2016learning,peng2019advantage}. Importantly, the suggested use of a value function instead of state-action value (aka a Q function) overcomes the lack of expert action annotations. In the second variant, we replace the 0-1 rewards with estimates from a learned discriminator as in~\citet{ho2016generative, zolna2020offline}. 
In both variants, we learn a value-function which effectively transfers expert knowledge from the unlabeled expert data onto the action labeled \background{} data. We name our approach Value learning from Observations (VfO).

In summary, we suggest the combination of IfO and iterative self-improvement in order to approach large scale behavior learning from a novel angle and under realistic, scalable data collection assumptions. To this end we introduce a new offline benchmark that is more representative of said setting. We further propose a novel algorithm (VfO), in two variants, that adapts offline RL mechanisms to imitation learning from observations. With a broad set of experiments, we confirm the representative power of our benchmark, underline the competitiveness of our algorithm, and show initial positive results of IfO applied to iterative self-improvement.

\section{Related Work}


The classical, straightforward approach to imitation learning is behaviour cloning (BC, \citet{osa2018algorithmic}), i.e., maximising the likelihood of actions in the dataset. However, this approach requires large numbers of optimal demonstrations. For this reason, a variety of methods have been developed to additionally benefit from suboptimal and other data sources by for instance extrapolating rewards from observations \citep{brown2019extrapolating, chen2021learning}, imitation via IRL \citep{ziebart2008maximum}, or even by using videos from generative models \citep{bharadhwaj2024gen2acthumanvideogeneration}. 

In online imitation learning, self-generated agent data represents the best data distribution to learn how to refine agent behaviour \citep{ross2011reduction, swamy2022minimax, lavington2022improved}. Different methods have been proposed to apply the reinforcement learning formalism to address an imitation problem - from classical and deep maximum entropy inverse RL \citep{ziebart2008maximum, wulfmeier2015maximum, barnes2024massively} to computationally more efficient adversarial imitation learning \citep{ho2016generative, fu2017learning, Wulfmeier2017MutualAT, kostrikov2018discriminatoractorcritic}. When treating imitation as matching of agent visited transitions or divergence minimisation, further divergences have been explored \citep{ke2021imitation, ghasemipour2020divergence}. 
Transitioning from adversarial learning to more stable value function optimisation, SQIL removes the intermediary classifier from GAIL and instead uses a binary reward \citep{reddy2020sqil}. IQlearn \citep{garg2021iqlearn} and ValueDICE \citep{Kostrikov2020Imitation} enable transitioning from explicitly defined to implicitly learned rewards.
Other non adversarial algorithms include PPIL \citep{viano2022proximal}, PWIL \citep{dadashi2021primal}, and CSIL \citep{watson2024coherent}. 
However, online imitation learning can be costly in domains like robotics, is sometimes not even possible and doesn't benefit from existing data sources.

When online data generation is impractical, suboptimal offline datasets can provide an alternative. 
Practical and scalable algorithms can be derived when using either discriminator \citep{zolna2020offline} or optimal transport \citep{luo2023optimal} based rewards together with offline RL.
Value \citep{kim2022demodice} and model-based approaches \citep{chang2021mitigating} expand the toolkit.
IQLearn can further be shown to be equivalent to BC with dynamics-aware regularisation term \citep{wulfmeier2024imitatinglanguagescalableinverse,sikchi2024dual}.

The online and offline settings described above require access to high-quality demonstrations with action annotations, often only attainable via complex tele-operation settings in robotics. 
The extension towards action-free demonstrations opens considerable scope and has been the target of further methods.
A separately trained inverse dynamics model can be applied to label action-free data, enabling behaviour cloning 
\citep{Radosavovic2021state,torabi2018generative}.
Learning rewards provides a path to instead relabel sub-optimal data for RL style optimisation \citep{eysenbach2021replacing, Davchev2021WishYW}.
Here, adversarial approaches present a common mechanism to learn rewards \citep{ho2016generative}. These can be adapted by controlling the discriminator input space, often benefiting from further regularisation \citep{zhu2020off, Liu2020State}.
Value function based imitation methods enabled by inverse Bellman updates and dual formulations of the problem like SMODICE \citep{pmlr-v162-ma22a} and DILO \citep{sikchi2024dual}, or variational formulations \citep{kostrikov2019imitation,garg2021iqlearn}, bypass the often hard to optimize adversarial objectives and are related to our approach.
While mathematically appealing, these methods can still be brittle and harder to scale to the real-world  directly from raw observations \citep{al2023ls,watson2024coherent}. Instead, we base our value-based algorithm on a simple RL backbone which draws on decades of experience.
We compare performance and show competitiveness against various baselines (including SMODICE and DILO).


\section{Method}
\label{sec:method}
\subsection{Offline Imitation Learning From Observations}
We consider learning in a dynamical system modelled as Markov decision process with states $s \in S$, actions $a \in A$, and dynamics $p(s_{t+1}|s_t, a_t)$.
In order to learn useful behaviour, the agent has access to two sources of information: a dataset of \expert{} state trajectories $\tau_E = (s_1, \ldots, s_T) \in D_E$ without actions and a dataset of possibly self-collected state-action trajectories of unspecified quality $\tau_B = (s_1, a_1, \ldots, s_T) \in D_B$. We refer to the latter as \background{} dataset. The agent's goal is to obtain a policy $\pi(a|s)$ that imitates the behaviour underlying the expert trajectories. To support scalability, we limit the use of further information (e.g. rewards, domain knowledge) and thus do not expect the agent to outperform expert performance.

Given the lack of \expert{} actions, the agent has to be able to leverage the \background{} dataset to understand the dynamics, i.e., the relationship between actions and states. However, similar to prior work on inverse RL \citep{Abbeel2004ApprenticeshipLV, ziebart2008maximum} as well as for the related problem of offline RL \citep{pmlr-v199-schweighofer22a, hong2023beyond}, the quality and distribution of the \background{} data plays an important role on the achievable performance. In previous work, different sources of \background{} data have been employed, such as agent replay data, a mixture of expert and non-expert data, or data collected with a suboptimal policy. Given that we are interested in an agent that can start from few assumptions and that should be able to leverage the data it collects, we focus on the suboptimal policy case. In order to generate a corresponding benchmark we suggest to train multiple policies using BC but vary the number of demonstrations provided. We then run these policies to collect multiple datasets of varying quality to form our self-improvement benchmark SIBench (see \Cref{sec:benchmarks}).

\subsection{VfO: Value from Observation}
\label{sec:vfo}

To tackle this novel setting we seek a simple method based on established learning mechanisms that can effectively learn from observations in the self-improvement setting. For this purpose, we consider two variants of a value-function based approach that learns a state-value function from observations alone.
In the first, we assign binary rewards (i.e., 1 for \expert{} and 0 for \background{}) to the data -- thus adapting SQIL~\citep{reddy2020sqil} to our setting. 
In the second, we use a learned discriminator that performs a soft \expert{} / \background{} assignment of each state -- thus adapting ORIL~\citep{zolna2020offline} to our setting. Other imitation learning-based rewards such as \citet{luo2023optimal} could also be employed, but note that employing rewards that rely on prior knowledge may impact generality (e.g. when derived from goal states).

\begin{algorithm}
\caption{Value from Observation (VfO)}\label{alg:cap}
\begin{algorithmic}
\REQUIRE Expert dataset $D_E$, background dataset $D_B$, mixture parameter $\alpha$, temperature $\lambda$, discount $\gamma$, initial policy $\pi_0$, initial value $v_0$. \textbf{Optional:} discriminator $d(s): S \mapsto [0, 1]$
\FOR{$k \gets 1$ to $K$ iterations}
    \STATE \color{blue}{$r(s', z) = \left\{\begin{matrix}  
      d(s') & \text{if discriminator provided}\\
      \mathbf{1}_E(z) & \text{otherwise} \\
      \end{matrix}\right.$}\color{black}
    \STATE $L_k^v \gets E_{(s, s', z) \sim \color{blue}(1-\alpha) D_E + \alpha D_B\color{black}} \newline
    ~~~~~~~~~~~ \left[\left(\gamma v_{k-1}(s') + r(s', z) - v_{k-1}(s)\right)^2\right]$
    \STATE $L_k^\pi \gets -E_{(s, s', a, z) \in D_B} \Big[\log(\pi_{k-1}(a|s)) \newline
    ~~~~~~~~~~~~ \exp\left((\gamma v_{k-1}(s') + r(s', z) - v_{k-1}(s))/\lambda\right)\Big]$
    \STATE $v_{k} \gets \text{AdamUpdate}(v_{k-1}, L_k^{v}),$
    \STATE $\pi_{k} \gets \text{AdamUpdate}(\pi_{k-1}, L_k^{\pi})$
\ENDFOR
\end{algorithmic}
\end{algorithm}

As mentioned above, we resort to learning a state-value function for transferring knowledge from the \expert{} data without action labels to the \background{} data. We note that a state-action Q-value based offline RL approach cannot be applied in our setting due to a lack of signal on the \background{} data: all transitions are labeled with a zero reward in the binary setting or potentially very small rewards in the learned discriminator setting. In contrast, if we apply an approach based on the state-value function $v$, policy evaluation is possible without knowing the action and can thus leverage a mixture of \expert{} and \background{} data.

First, we define a virtual policy $\bar\pi$ which mixes the \expert{} and \background{} data-generating processes at each transition:
\begin{align}
    \bar\pi(a|s) = p(z=E|s, \alpha) \pi_E(a|s) + p(z=B|s, \alpha) \pi_B(a|s)
\end{align}
with mixture coefficient $\alpha$ and where $z$ denotes the latent indicating the origin of the data, either \expert{} $E$ or \background{} $B$.
This policy is equivalent to deciding at the beginning of an episode whether to follow the implicit expert or background policy.
Note that the probability of using $\pi_E$ or $\pi_B$ is state-dependent, and will depend on the likelihood of reaching $s$ under each policy.
With discount factor $\gamma$ and reward $r(s', z)$, we can define the temporal difference error of a value function for this policy:
\begin{align}
    L_v = E_{(s, s', z) \sim (1-\alpha) D_E + \alpha D_B} (\gamma v(s') + r(s', z) - v(s))^2.
\end{align}

Leaving aside -- for a moment -- how the reward can be obtained, we can utilize this learned value function to find an improved policy by (exponentiated) advantage weighted regression \citep{peng2019advantage,wang2018exponentially}. This amounts to weighted supervised learning on the \background{} data (which enforces closeness to the policy that generated the data via a temperature $\lambda$) and yields the following loss:
\begin{align}
    L_\pi = -E_{(s, s', a) \in D_B} [&\exp((\gamma v(s') + r(s', z) - v(s))/\lambda) \nonumber \\
    &\log(\pi(a|s))].
\end{align}

An advantage of this presented scheme is its simplicity and use of well established offline RL methods, allowing for an efficient implementation while utilizing insights from many years of RL research such as the use of target networks and how to deal with terminations (see \cref{sec:exp}). The full method is described in \Cref{alg:cap} with the key differences to the offline RL setting marked in blue: the reward source and the mixture of \expert{} and \background{} data.

\paragraph{Binary demonstration-based rewards (VfO-bin)}
In the simplest setting we avoid any additional learning or estimation bias in the reward function by directly assigning a reward of 1 to \expert{} transitions and a reward of 0 to \background{} transitions in line with what has been proposed by \citet{reddy2020sqil}. While this might seem trivial, it recovers what a perfect discriminator with infinite capacity would output and removes a layer of complexity. It follows the intuition that we may be able to leverage the value function directly for distinguishing good from bad states in the \background{} data, rather than learning an intermediary reward function. We can make this notion more precise by realising that when we only provide a reward of $1$ for the \expert{} transitions and $0$ otherwise, the learned value can be interpreted as:
\begin{align}
    v_{\bar\pi}(s_t) &= p(z_t=E|s_t) \nonumber \\
    & \quad + \gamma E_{a \sim \bar\pi(\cdot|s_t), s_{t+1} \sim p(\cdot|s_t, a)} v_{\bar\pi}(s_{t+1}), \nonumber \\
    &= E_{(s_{t+1}, s_{t+2}, \ldots) \sim \bar\pi} \sum_{i=0} \gamma^i p(z_{t+i}=E|s_{t+i}),
\end{align}
which is the cumulative discounted likelihood of futures states having been visited by the expert within the mixed dataset when starting in $s_t$. A policy that maximizes this cumulative return thus prioritises visiting expert states.

A different way to look at it, is that states that are contained in both the \expert{} and \background{} data will receive a positive and a negative reward signal. Given that policy improvement relies on weighted regression on the \background{} data this is perfectly fine: actions that lead to states which are closer to the \expert{} data will receive a higher weight. 

\paragraph{Discriminator-based rewards (VfO-disc)}
In a second setting, we consider learning a discriminator to serve as reward~\citep{ho2016generative}. When learning from observations, these usually learn to distinguish expert from non-expert states~\citep{zolna2020offline,pmlr-v162-ma22a} in order to derive a reward for learning a policy. 
We adopt the objective from ORIL \citep{zolna2020offline} and pre-train the discriminator by minimizing 
\begin{equation}
    L_d = E_{s \sim D_E} [-\log d(s)] + E_{s \sim D_B} [- \log (1 - d(s))],
\end{equation}
where $d(s) \in [0, 1]$ is a binary classifier and the objective is akin to training a discriminator in generative adversarial learning \citep{goodfellowGAN,ho2016generative}. The discriminator output directly serves as reward similar to what was done in \citet{Wulfmeier2017MutualAT} using the Wasserstein-1 distance \citep{pmlr-v70-arjovsky17a}. 

\section{Experiments}
\label{sec:exp}

\begin{figure*}[ht]
    \centering
    \includegraphics[width=.3\linewidth,trim={2cm 1cm 0cm 2cm},clip]{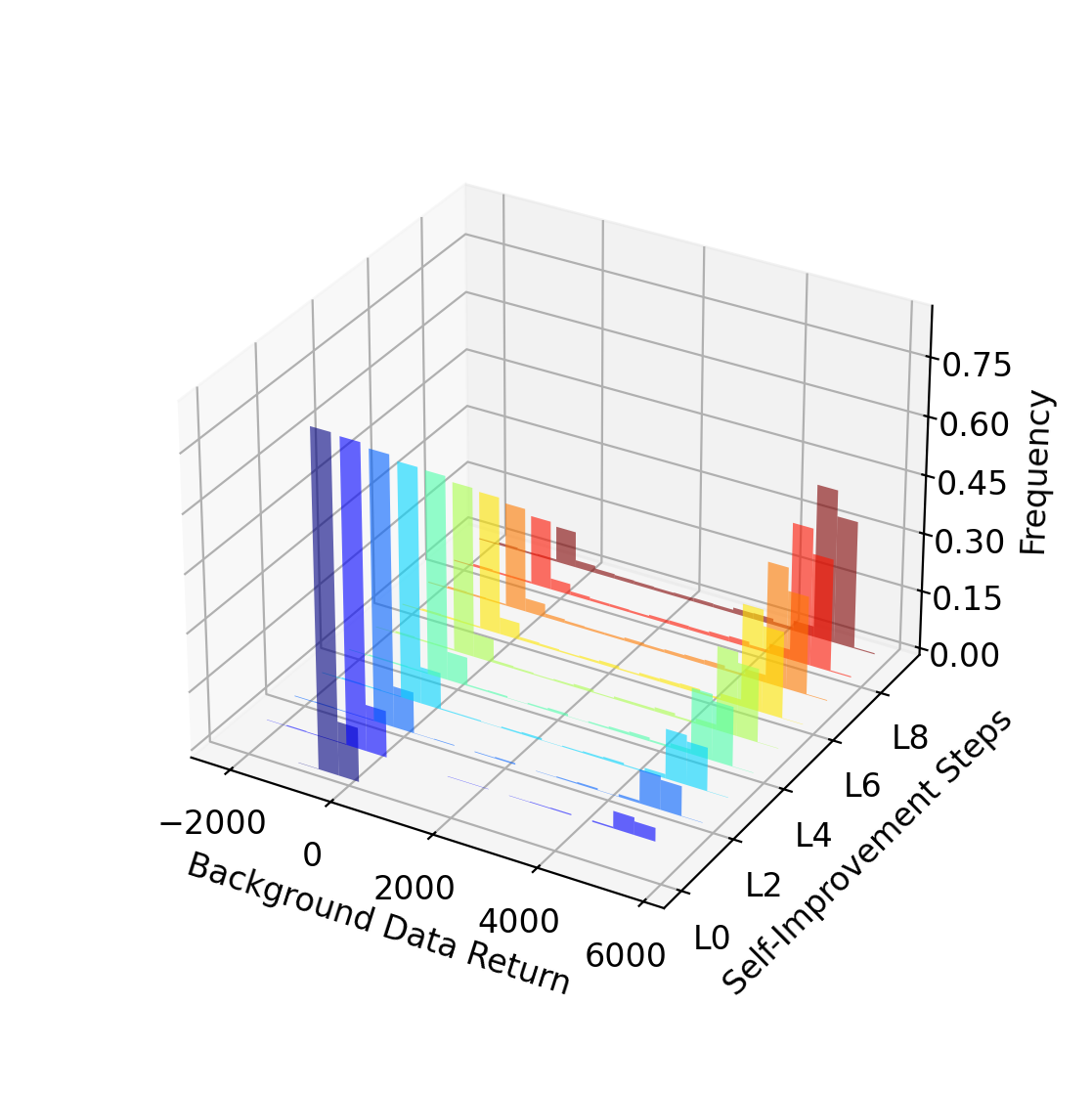}
    \includegraphics[width=.3\linewidth,trim={2cm 1cm 0cm 2cm},clip]{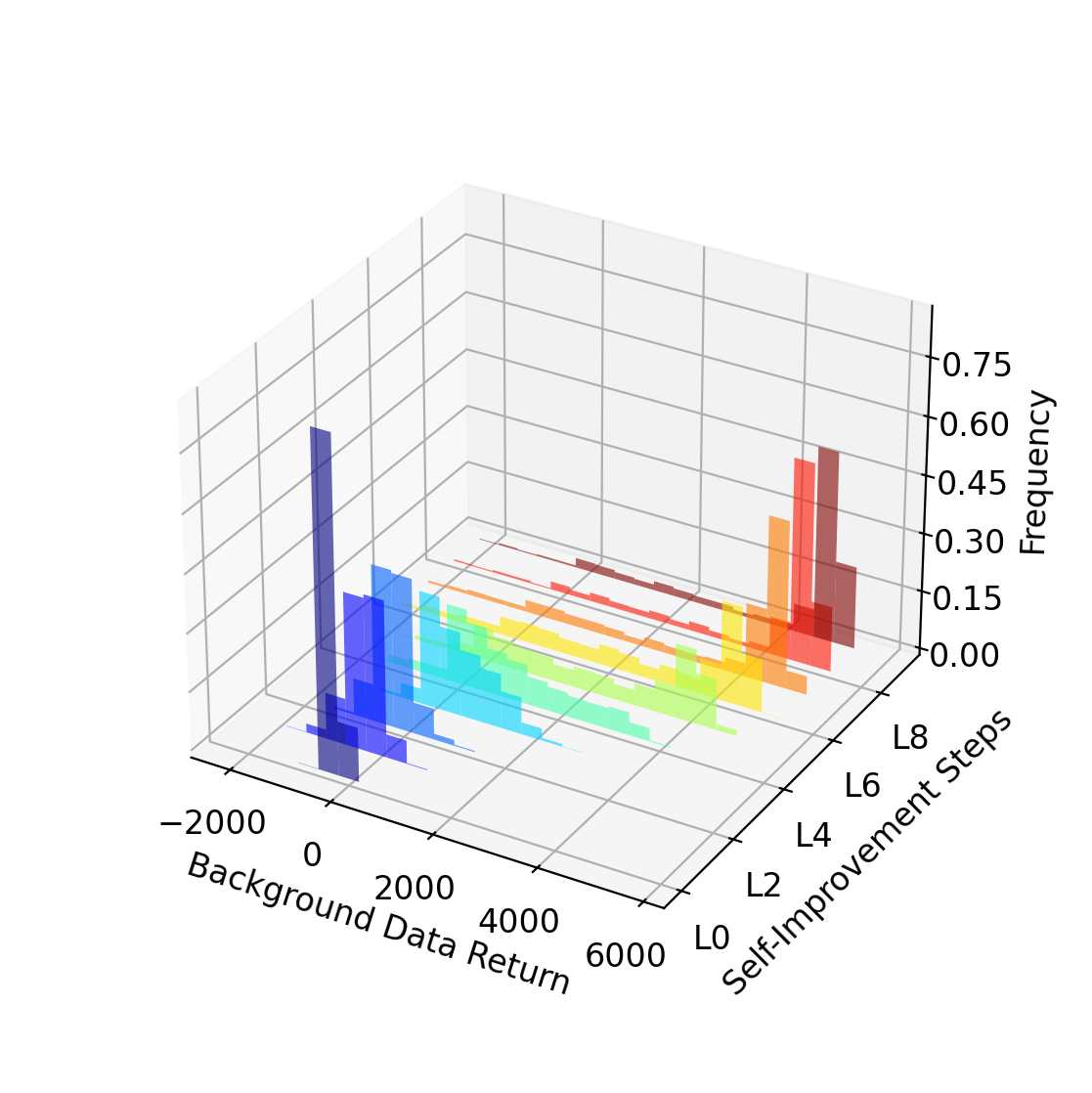}
    \includegraphics[width=.3\linewidth,trim={2cm 1cm 0cm 2cm},clip]{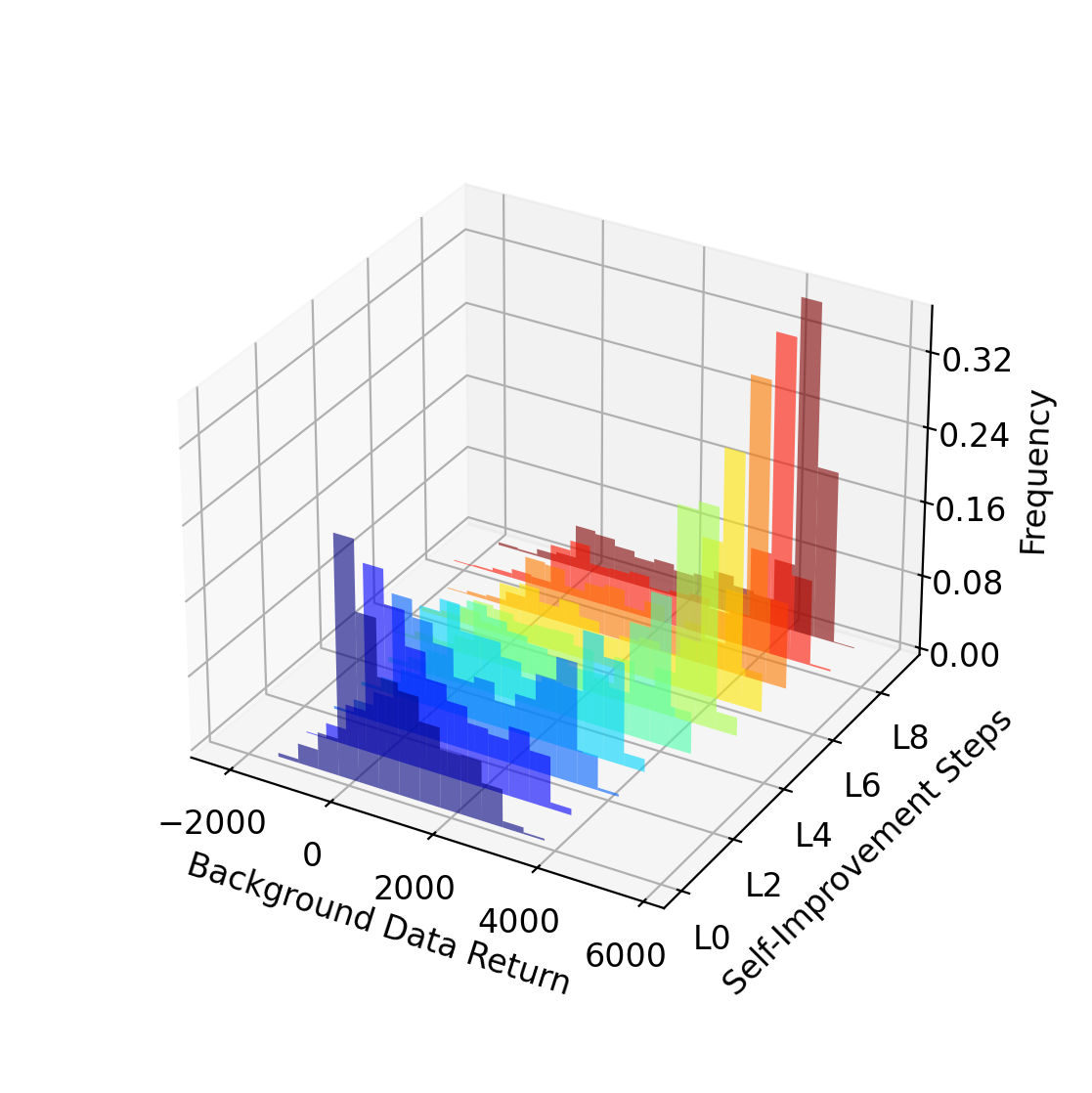}\\
    \scriptsize \hspace{30mm} Bimodal\hfill Self-Improvement Benchmark \hfill Self-Improvement \hspace*{20mm} \\
    \caption{\emph{Background} data return distributions for different data configurations and different quality levels (L0-L9). Comparing bimodal (mixing expert and random policy data), self-improvement benchmark (data originating from single policies of different quality), and self-improvement (exemplary data encountered during iterative self-improvement) intuits how different algorithmic properties benefit in each configuration. Explicitly learning discriminators between demonstrations and other data is intuitively easier with stronger splits (bimodal). The intermediate offline self-improvement benchmark serves as a good proxy and enables quick evaluation bypassing serial dependencies between self-improvement steps. \label{fig:data}}
\end{figure*}

We aim to compare different methods across different data distributions. In particular we want to investigate the difference between the bimodal distribution with a clear gap in the \background{} data and our suggested self-improvement benchmark, with a more nuanced distribution without a clear expert mode (both illustrated in \Cref{fig:data}).
While the bimodal setting is commonly employed for evaluation in previous work we argue that our self-improvement benchmark can better serve as an offline proxy for the iterative self-improvement that motivates our paper.
The following questions guide our experiments:
\begin{itemize}
    \item How do different data distributions affect the performance across algorithms and how effective is VfO? (\Cref{sec:sibench,sec:bimodal})?
    \item Does the offline self-improvement benchmark correlate with full iterative self-improvement (\Cref{sec:self-improvement})?
    \item Can VfO deal with complex inputs such as images (\Cref{sec:vision})?
\end{itemize}

\subsection{Datasets and Benchmarks} \label{sec:benchmarks}
Our experiments are based on two established benchmarks: D4RL \citep{fu2021d4rl} with many baseline results in the literature and Robomimic \citep{robomimic2021}, from which the robosuite \citep{robosuite2020} tasks in particular provide a more realistic challenge for our algorithms -- and include tasks that require the ability to process image data. 
D4RL contains data from various OpenAI gym \citep{openaigym} MuJoCo \citep{todorov2012mujoco} environments, from which we use the Ant, HalfCheetah, Hopper and Walker2D domains. The expert data for these environments comes from policies trained via RL. For each domain, there are 1000 expert demonstrations available; and we drop the action information to obtain the IfO setting. From the simulated robosuite domains, we use Lift, PickPlaceCan, and NutAssembly. For each of these tasks, there are 200 human demonstrations.

The most realistic, scalable source of \background{} robot data is agent-generated. Therefore in contrast to existing \background{} datasets -- which are mostly bimodal (very high and low performance) -- we target self-improvement as a data source. To emulate the sequential nature of different quality levels that we would expect during self-improvement, we introduce a proxy Self-Improvement Benchmark (SIBench) which uses data generated by a set of varying policies:
For each task $\tau$ we train a set of BC policies $\{\pi_{\tau}^d | d \in \{1, 2, 5, 10, 20, 50, 100, 200, 500, 1000\}, d\leq N_\tau\}$, where $d$ indicates the number of demonstrations that the policy was trained on and $N_\tau$ is the number of demonstrations available. 
For each of these policies, we collect 1000 episodes to form distinct \background{} datasets. This setting enables considerably faster experimental iteration compared to running a full self-improvement experiment (since the data is pre-generated and fixed) and creates a consistent benchmark for fair comparisons; but comes at the cost of removing the data generation from the analysis. 
Importantly, looking at the return distributions \Cref{fig:data}, we can observe that this proposed \background{} data is qualitatively close to the data encountered in self-improvement.
Further, the different \background{} datasets will exhibit different levels of overlap with the \expert{} data, ranging from a scenario where the \expert{} data is mostly out of distribution to a regime where the \expert{} data is contained in the \background{} data.

For comparability with prior work and to investigate the added value of our evaluation scheme, we also constructed datasets of what we refer to as bimodal data composed of expert demonstrations and trajectories generated with a random policy (i.e., actions sampled from a uniform distribution). To obtain a more complete picture we sweep over the data mixture: we interpolate linearly from 1000 random demonstrations to 1000 expert demonstration.

Using 1000 simulated rollouts of the stochastic policy, we compute average returns and success rates\footnote{Note, that this may not be the best metric for imitation learning performance, as it may provide a distorted view of imitation. E.g. success rate may be blind to any improvement as long as the task is not solved.} and report mean and standard deviation across 5 seeds. We display most results as the difference between the average return obtained from the learned policy and the average return observed in the \background{} data. In contrast to simply reporting policy returns, we argue that this clearly visualises improvement of imitation learning algorithms and allows for better comparison against baselines via improved resolution.

\begin{figure*}[ht]
    \centering
    \includegraphics[width=.22\linewidth]{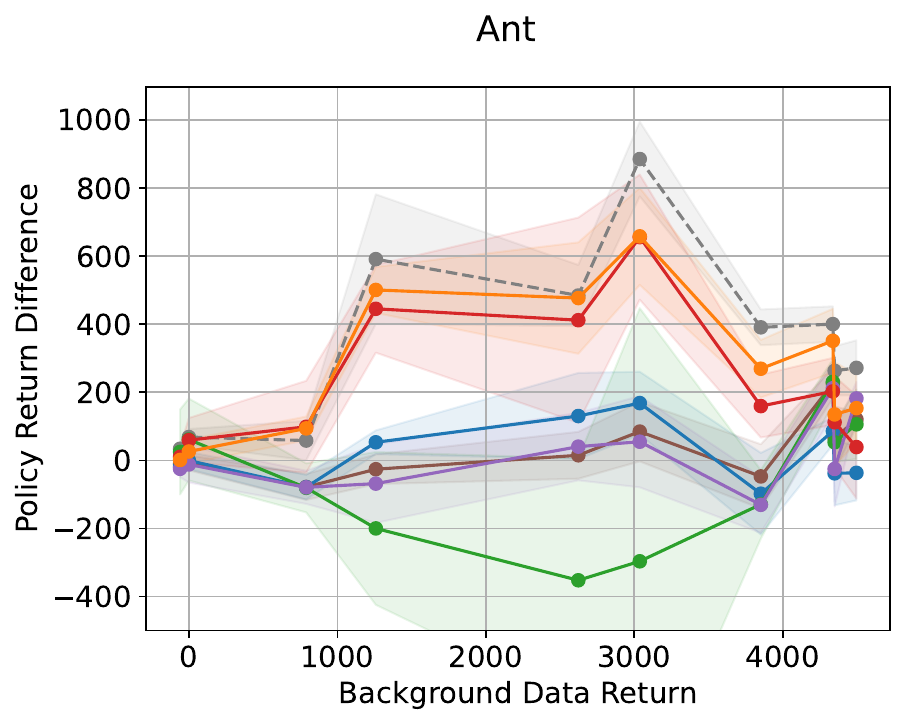}
    \includegraphics[width=.22\linewidth]{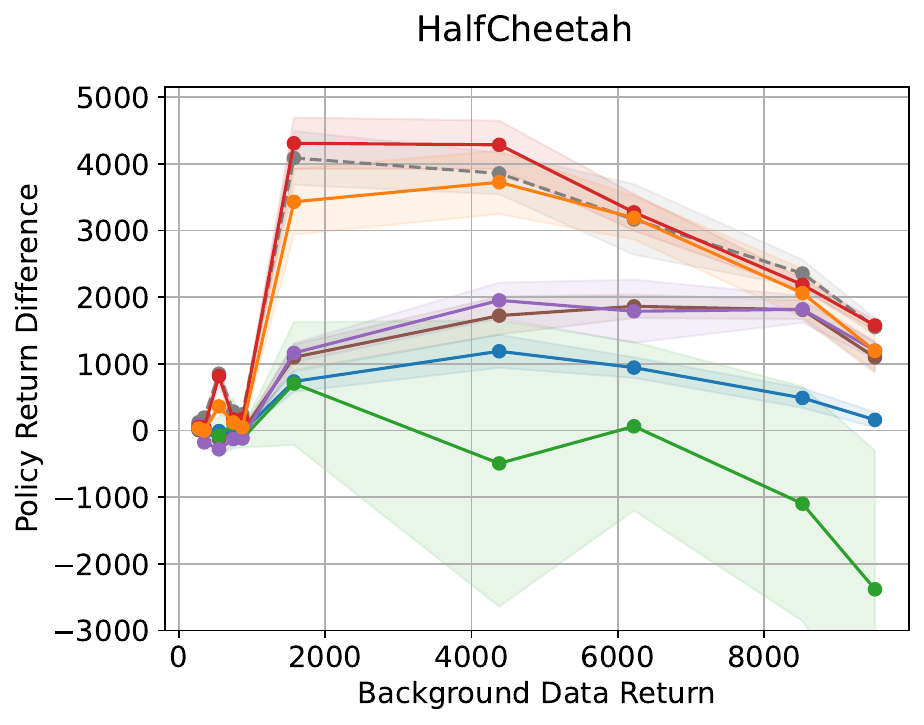}
    \includegraphics[width=.22\linewidth]{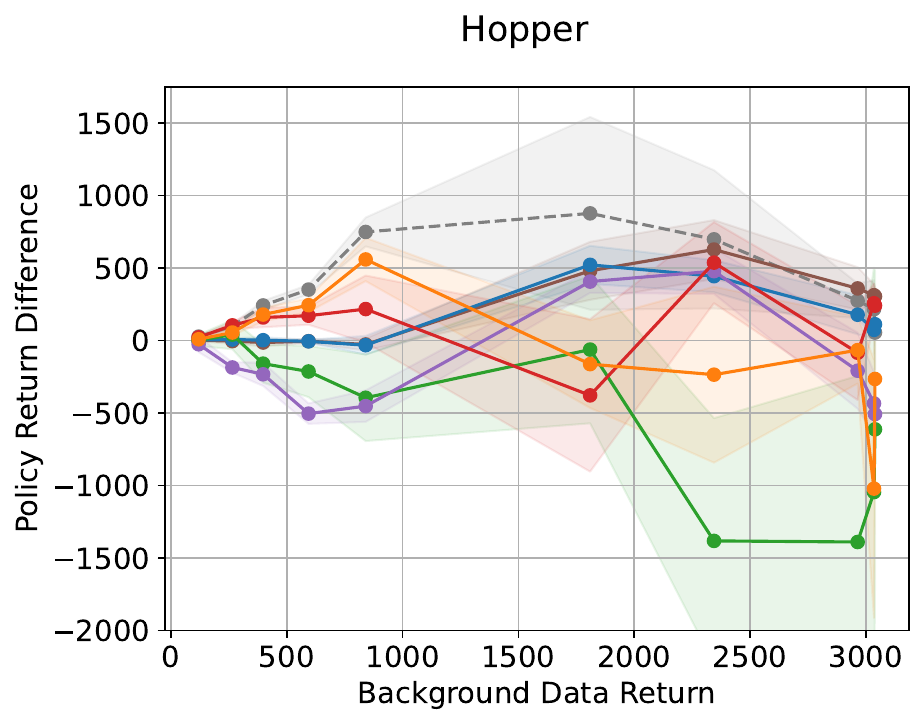}
    \includegraphics[width=.22\linewidth]{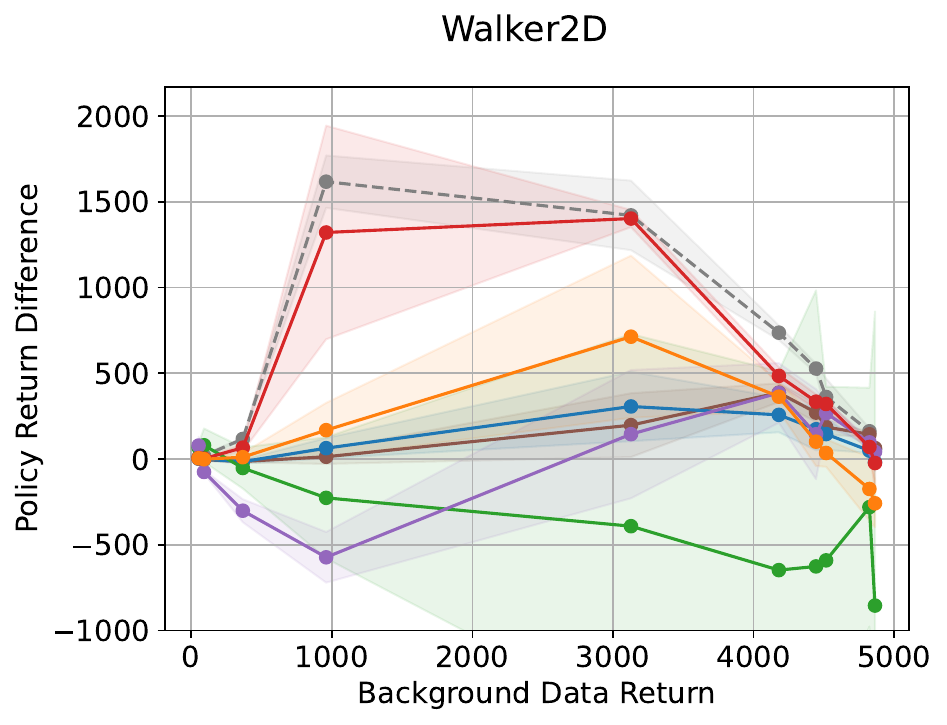}
    \includegraphics[width=0.09\linewidth,trim={9cm -1.5cm 8cm 3cm},clip]{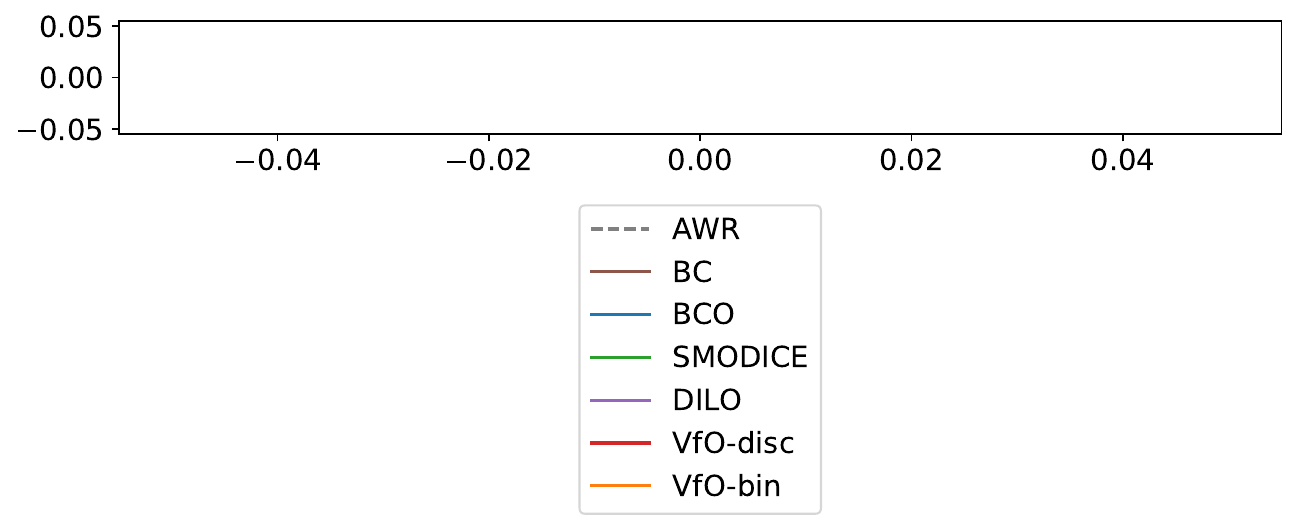}
    \caption{Difference in cumulative return of various algorithms on D4RL tasks using the SIBench data. We plot results against the average return in the background data. Positive differences mean the policy produced by the algorithm is better than the policy that generated the background data.
    VfO-disc and VfO-bin show good improvement across the spectrum of \background{} data and are comparable with the oracle AWR. SMODICE and DILO only rarely improve on the data.}
    \label{fig:d4rl_irlb}
\end{figure*}

\begin{figure*}[ht]
    \centering
    \includegraphics[width=.24\linewidth]{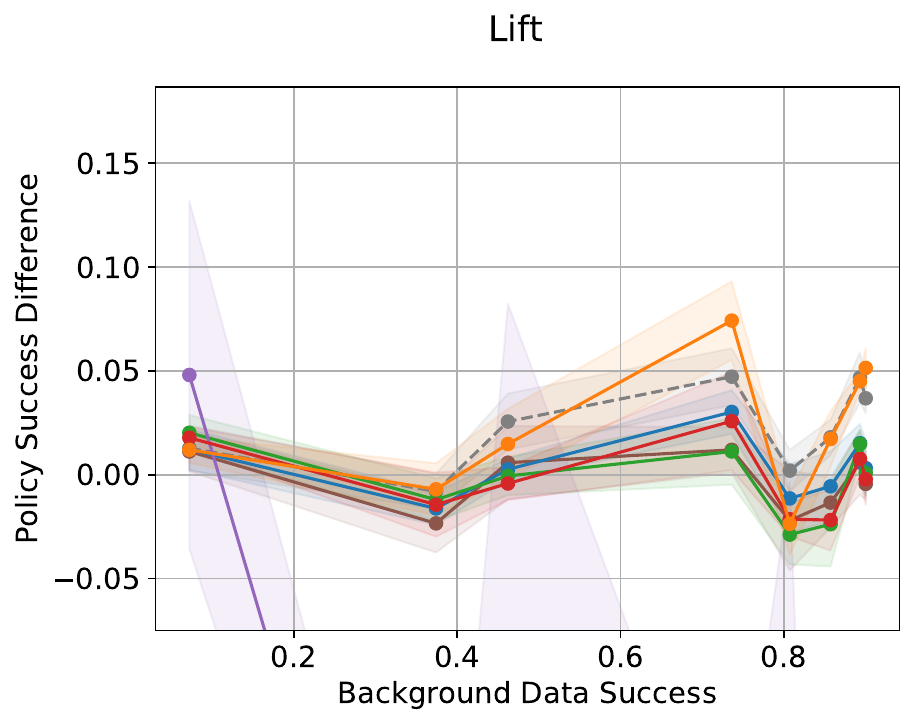}
    \includegraphics[width=.24\linewidth]{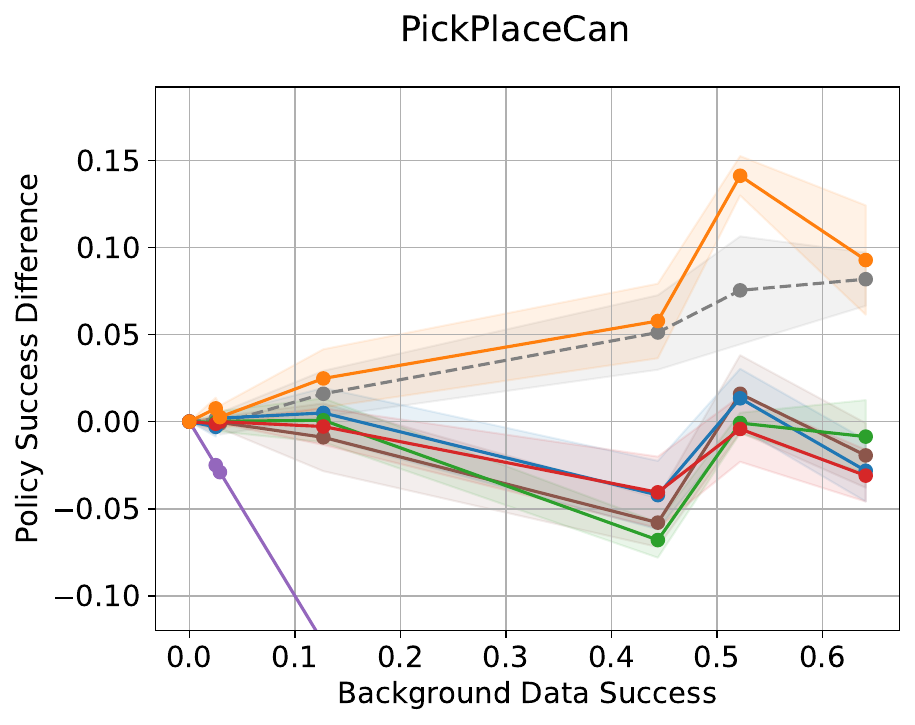}
    \includegraphics[width=.24\linewidth]{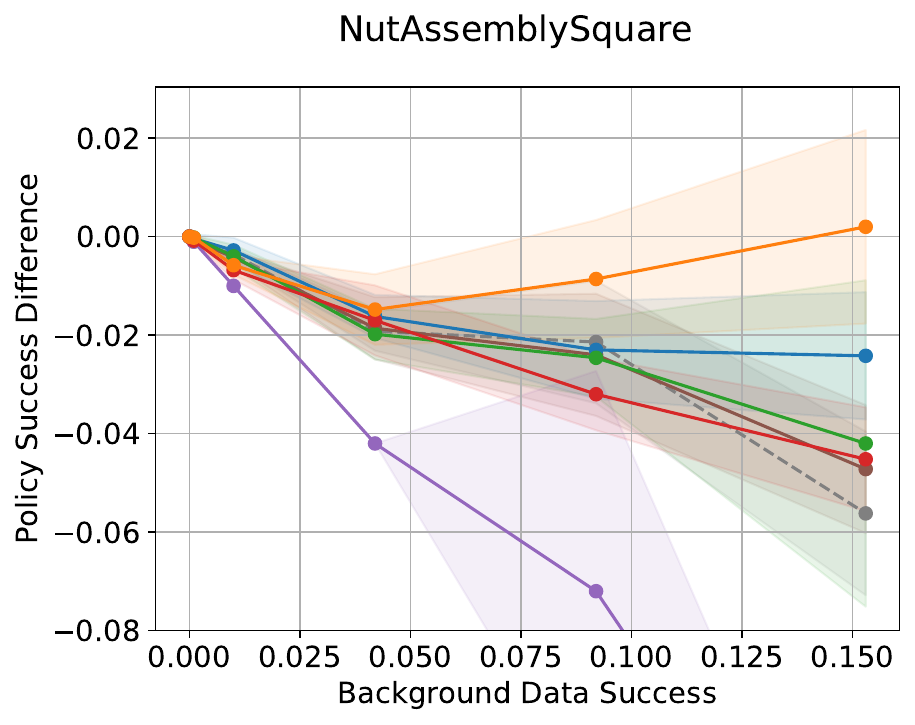}
    \includegraphics[width=0.09\linewidth,trim={9cm -1.5cm 8cm 3cm},clip]{figures/legend_v.pdf}
    \caption{Difference in success of various algorithms on Robomimic tasks using the SIBench data. VfO-bin and the oracle AWR mostly yield improvement. DILO, SMODICE, and VfO-disc do not exhibit improvement.}
    \label{fig:robomimic_irlb}
\end{figure*}

\subsection{Algorithms and Baselines}
We compare VfO to a broad set of baselines including behaviour cloning (BC) on the \background{} data, BCO \citep{torabi2018behavioral}, SMODICE \citep{pmlr-v162-ma22a}, and DILO \citep{sikchi2024dual}\footnote{Re-implemented in a shared code base for improved comparability. See \Cref{app:baselines} for details. Where possible we verify performance against reported results in prior work.}. 
Finally, to provide a performance upper bound and thereby support usefulness of the \background{} data, we also report results for Advantage-Weighted Regression (AWR; \citealp{peng2019advantage}) trained with ground-truth rewards on the \background{} data. This can be interpreted as an oracle algorithm that does not perform IfO (and is not directly comparable) but serves as indicator of what could be learned from the data. Please refer to \Cref{app:implementation,app:baselines} for further implementation details.


\subsection{SIBench Results}
\label{sec:sibench}

The results for the SIBench D4RL tasks are shown in \Cref{fig:d4rl_irlb} (see plots with absolute returns in \Cref{app:abolute} and SQIL with privileged actions in \Cref{app:asqil}). Both VfO-bin and VfO-disc perform well on the Ant and HalfCheetah tasks, getting close to the oracle AWR performance across the full spectrum of \background{} data. All methods underperform on the Hopper task. On Walker2D, VfO-disc performs on par with AWR and better than VfO-bin. A possible explanation for VfO-bin's decreased performance could be its lack of immediate reward on the \background{} data which could impact its performance on cyclic tasks. SMODICE and DILO perform poorly in comparison on all tasks, only improving on the data on scattered occasions. Generally in this more realistic settings VfO performs remarkably strong; close to AWR's oracle performance in many settings despite it having to deal with lack of reward information and lack of action data on the expert demonstrations. Regarding the performance of SMODICE and DILO, we hypothesize that similar to residual gradient algorithms in RL \citep{baird1995residual}, which do not use stop-gradients or target networks, the signal from the Bellman residual may be very weak when there is significant overlap between good and bad trajectories, such as is the case for the \background{} data here.

\begin{figure*}[ht]
    \centering
    \includegraphics[width=.22\linewidth]{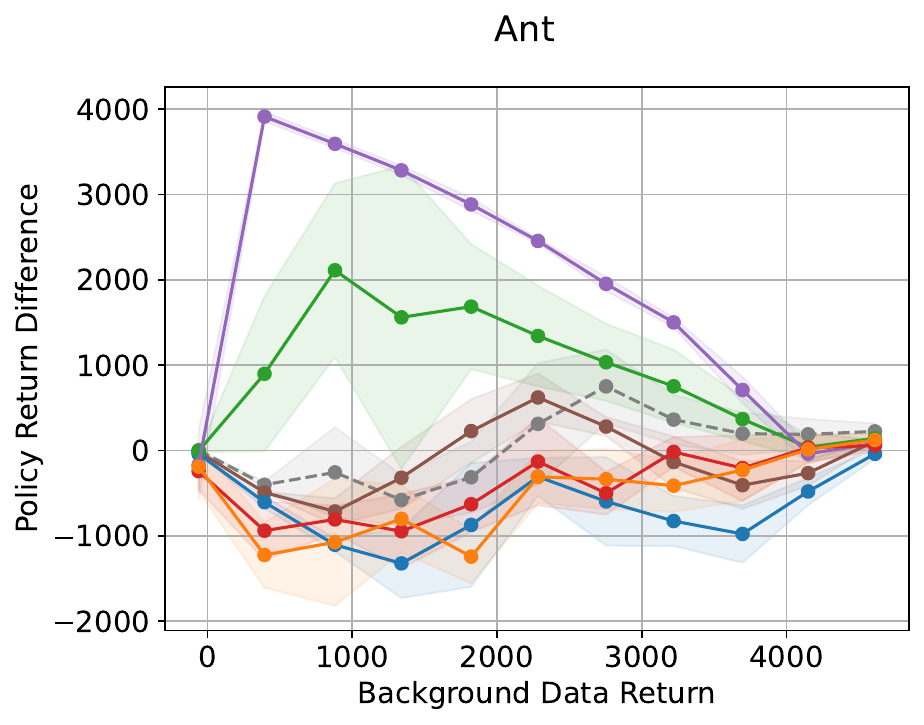}
    \includegraphics[width=.22\linewidth]{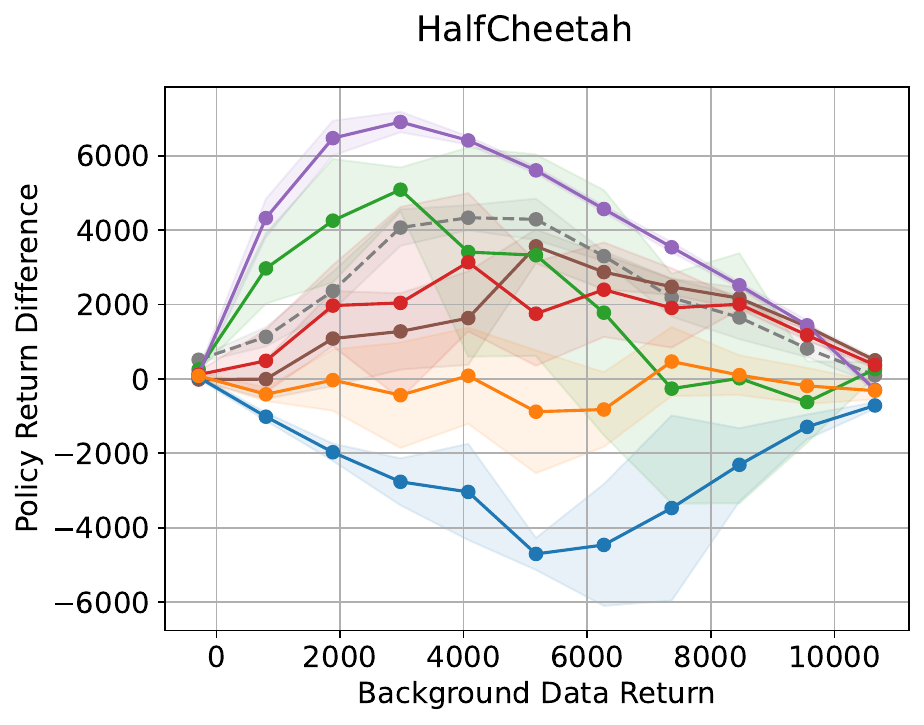}
    \includegraphics[width=.22\linewidth]{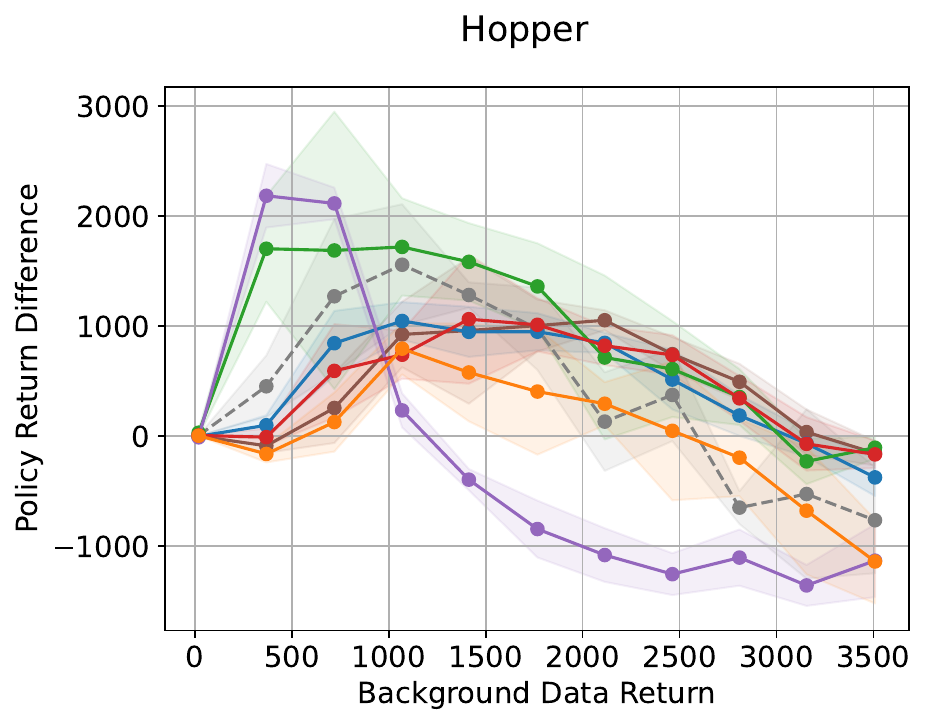}
    \includegraphics[width=.22\linewidth]{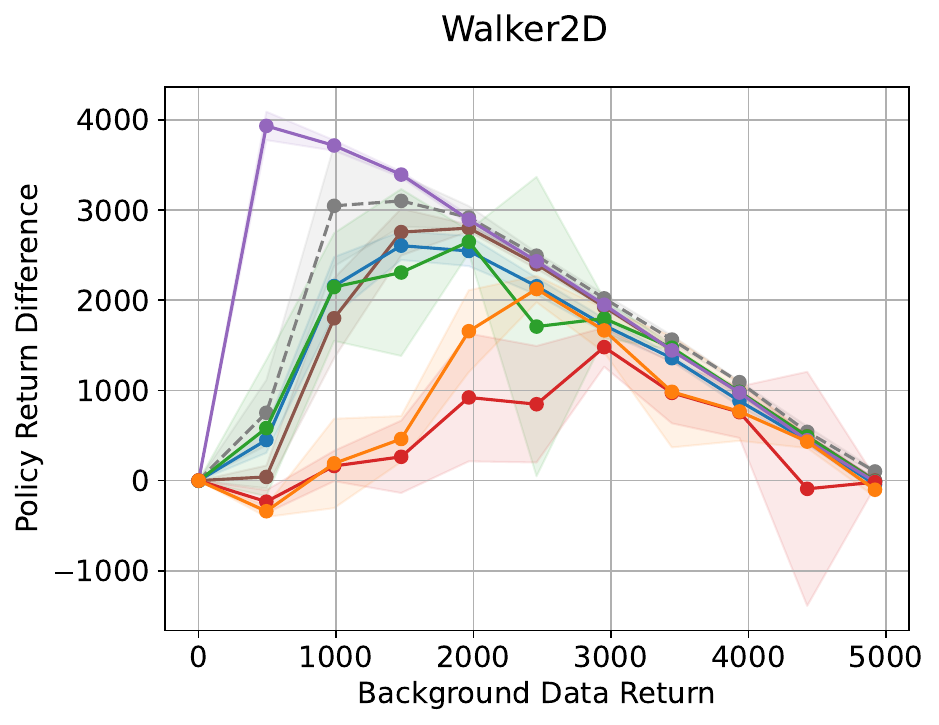}
    \includegraphics[width=0.09\linewidth,trim={9cm -1.5cm 8cm 3cm},clip]{figures/legend_v.pdf}
    \caption{Difference in cumulative return of various algorithms on D4RL tasks using the bimodal data. As reported in previous work, SMODICE and DILO exhibit strong improvement when the data is composed of a little amount of expert demonstrations. VfO-bin and VfO-disc both underperform in that case.}
    \label{fig:d4rl_bimodal}
\end{figure*}

\begin{figure*}[ht]
    \centering
    \includegraphics[width=.24\linewidth]{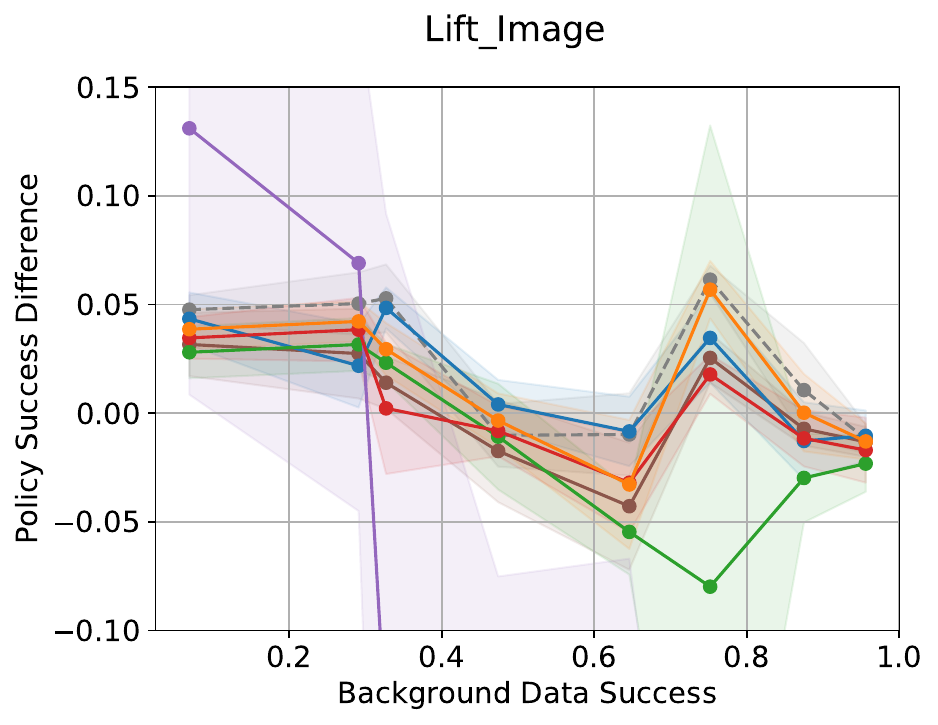}
    \includegraphics[width=.24\linewidth]{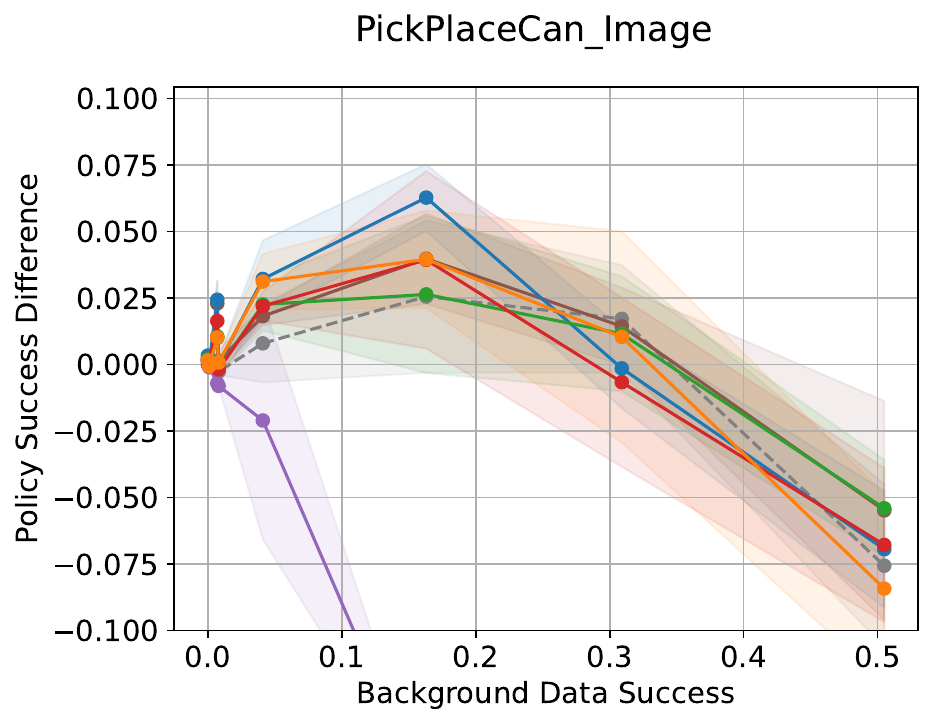}
    \includegraphics[width=.24\linewidth]{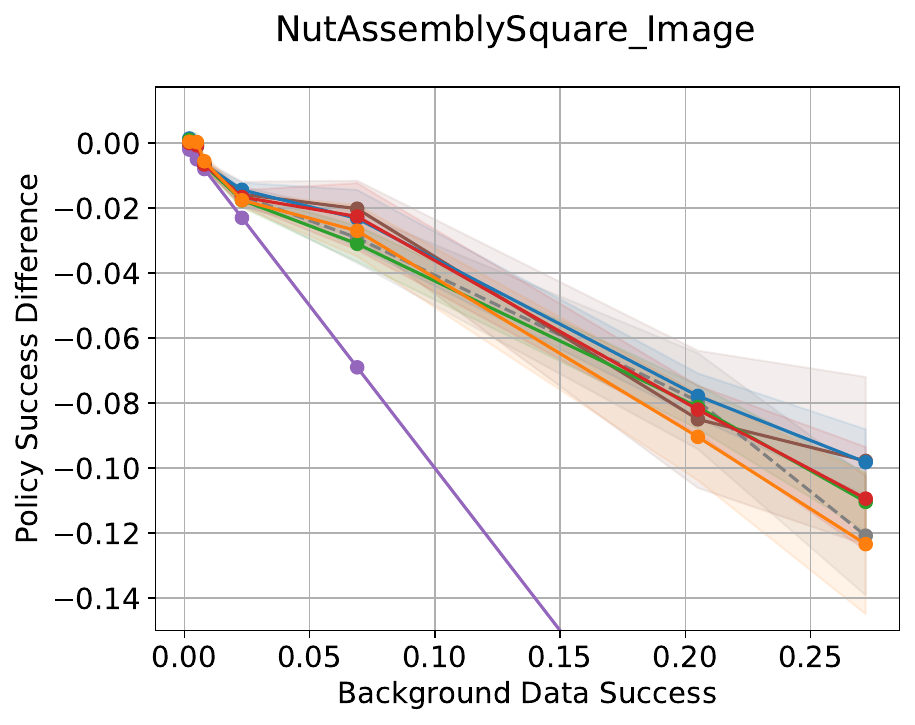}
    \includegraphics[width=0.09\linewidth,trim={9cm -1.5cm 8cm 3cm},clip]{figures/legend_v.pdf}
    \caption{Difference in success of various algorithms on Robomimic tasks using the SIBench image data. In this difficult setup, with images and human operator based demonstrations, VfO-bin and AWR manage to achieve some improvement on Lift.}
    \label{fig:robomimic_image_irlb}
\end{figure*}

\Cref{fig:robomimic_irlb} depicts the results for the SIBench Robomimic tasks. Overall the results are less conclusive here and this could be related to how improvement is measured: Given that there is no dense reward for these tasks, we resort to success, which is less indicative of learning progress, i.e. behavior could become more similar to the expert demonstration without higher success rate.
Further, demonstrations are generated by human operators which could further affect performance \citep{robomimic2021}.
Nevertheless, VfO-bin is able to yield positive improvement across most tasks and performs on par or better than AWR, while VfO-disc, SMODICE, and DILO perform worse.
Comparing AWR against VfO effectively also compares the underlying driving sources of information, i.e. reward annotations against demonstration. In a scenario with sparse rewards, such as for Robomimic, it is entirely possible that relying on demonstrations allows for better performance.

\subsection{Bimodal Results}
\label{sec:bimodal}

To investigate how SIBench data differs from existing data mixtures and to enable comparability with previous work, we report results for the bimodal D4RL data in \Cref{fig:d4rl_bimodal}. DILO and SMODICE exhibit strong improvement, particularly when the data is composed of relatively few expert demonstrations. In most cases, DILO reaches expert performance for the third data point (200 expert demonstrations) which is in line with results from \citet{sikchi2024dual}. Improvement then degrades smoothly with increasing data quality as there is less room for improvement.

VfO-bin and VfO-disc underperform when compared to BC. The improvement of BC itself can be attributed to the bimodal state distribution and to dynamic effects that can lead BC to pick the more consistent underlying policy \citep{zhangtranscendence}. These results are also in line with our hypothesis that bimodal data of this type turns imitation learning into a filtering problem of separating the good from bad trajectories and may not effectively measure the properties of imitation learning algorithms that matter in practical self-improvement scenarios. Instead, the mostly distinct \expert{} and \background{} state distributions render it more important to pick the right action when the distributions bifurcate. VfO likely struggles to do so because the learned values are not sufficiently discriminative. However, lowering temperatures to increase the effect incurs instabilities.

\subsection{Vision-based Results}
\label{sec:vision}
We also investigate the ability to learn from image observations (see \Cref{fig:robomimic_image_irlb}), as is often required in real-world robotics applications. For this we report results on the Robomimic tasks using the SIBench image data. While improvement is more difficult to measure here, we can observe some improvement in the Lift domain for AWR and VfO-bin, hinting that our simple VfO scheme is a strong candidate even in high-dimensional, difficult settings.

\subsection{Iterative Self-Improvement Results}
\label{sec:self-improvement}

\begin{figure*}[ht]
    \centering
    \includegraphics[width=.24\linewidth]{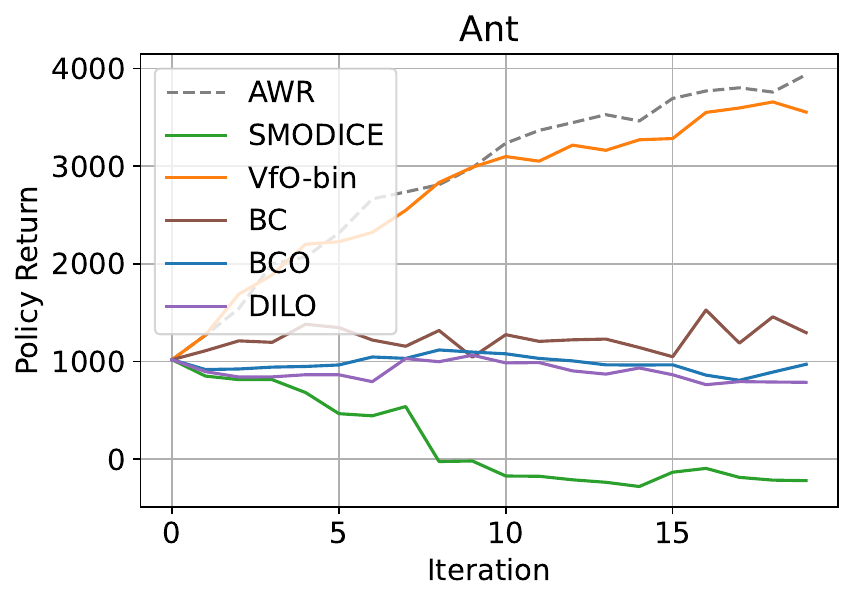}
    \includegraphics[width=.24\linewidth]{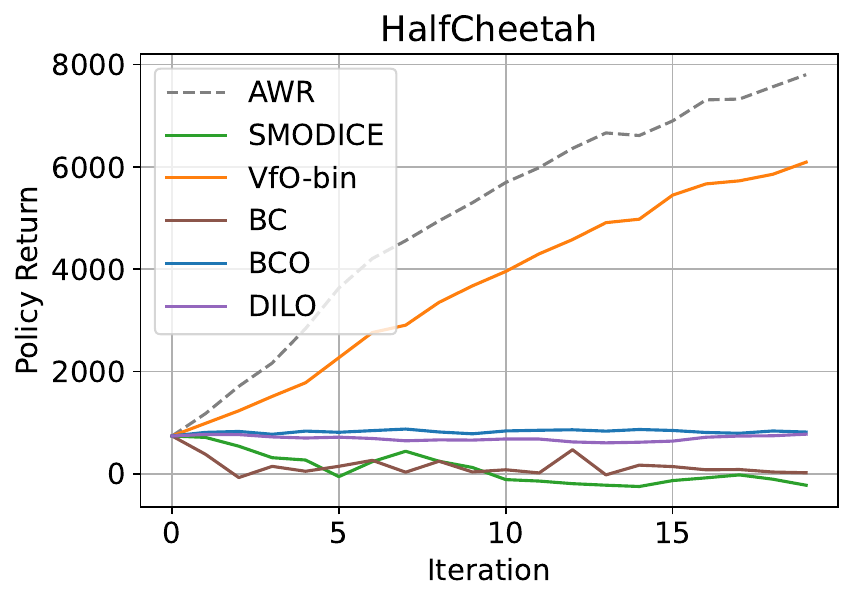}
    \includegraphics[width=.24\linewidth]{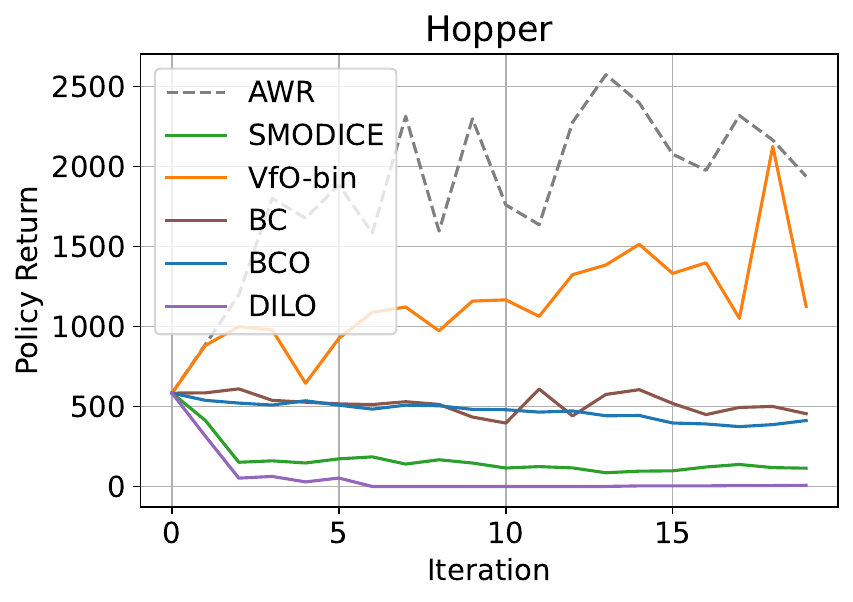}
    \includegraphics[width=.24\linewidth]{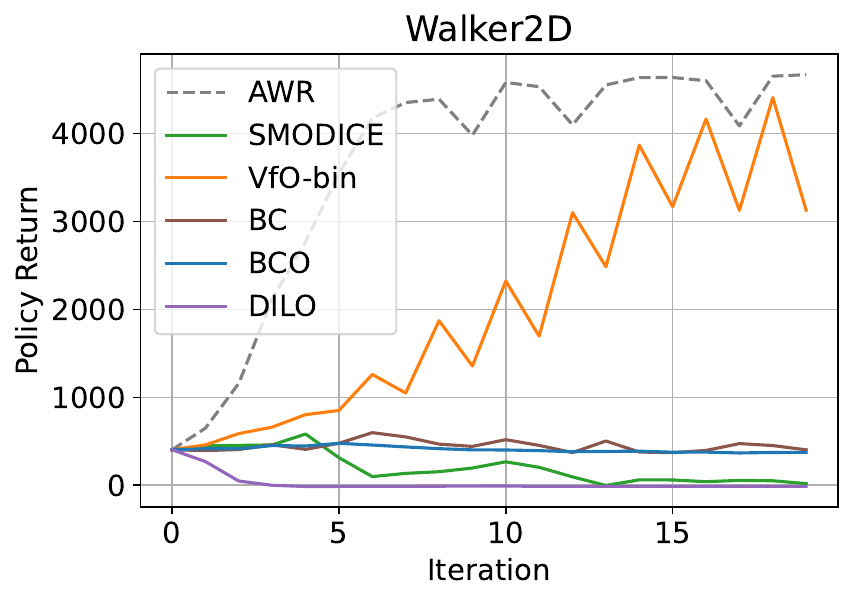}
    \caption{Self-improvement experiments for D4RL tasks. We evaluate VfO-bin, BC, BCO, SMODICE, DILO, and AWR with ground-truth rewards. Starting with low-performance initial policies, we generate data to train the next iteration of policies for each algorithm and iterate. The average policy return is plotted for each iteration. Both AWR with ground-truth rewards and VfO-bin lead to strong results.}
    \label{fig:d4rl_zth}
\end{figure*}

\begin{figure*}[ht]
    \centering
    \includegraphics[width=.24\linewidth]{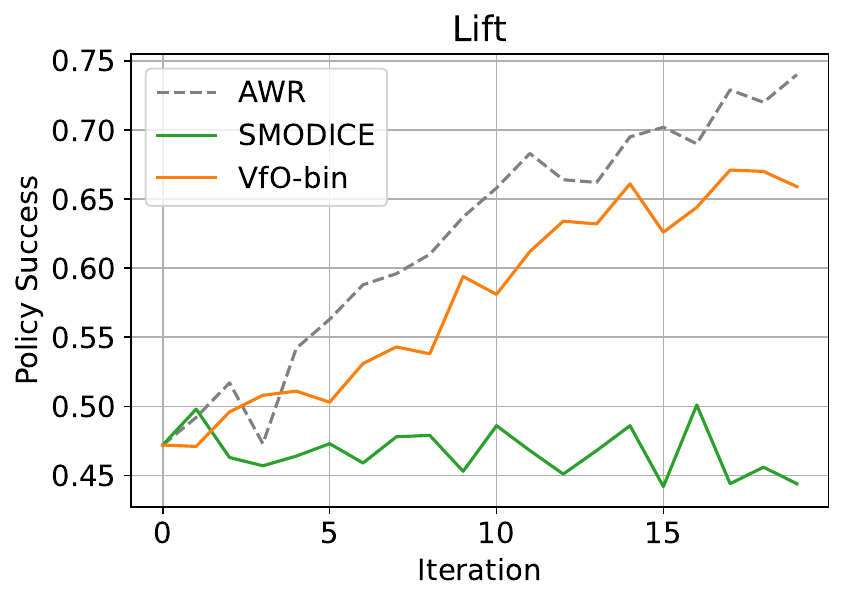}
    \includegraphics[width=.24\linewidth]{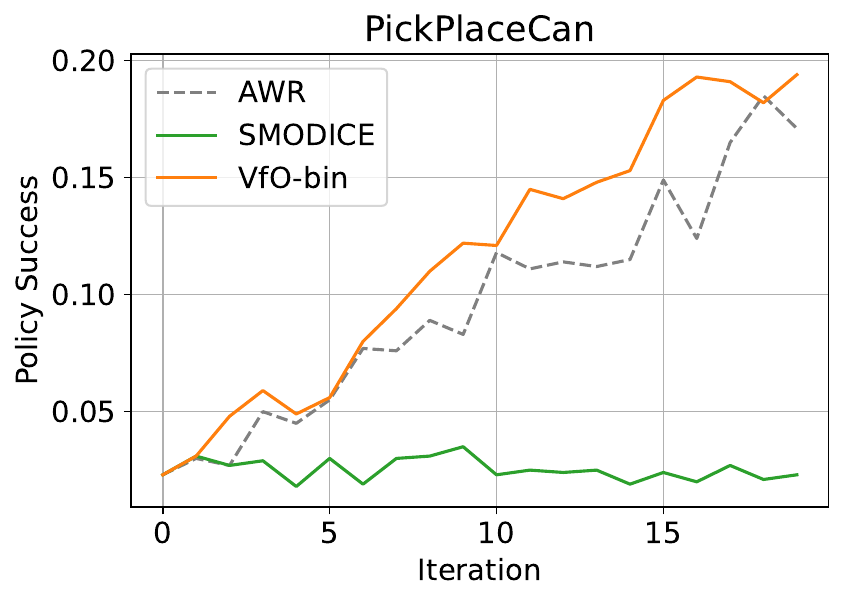}
    \includegraphics[width=.24\linewidth]{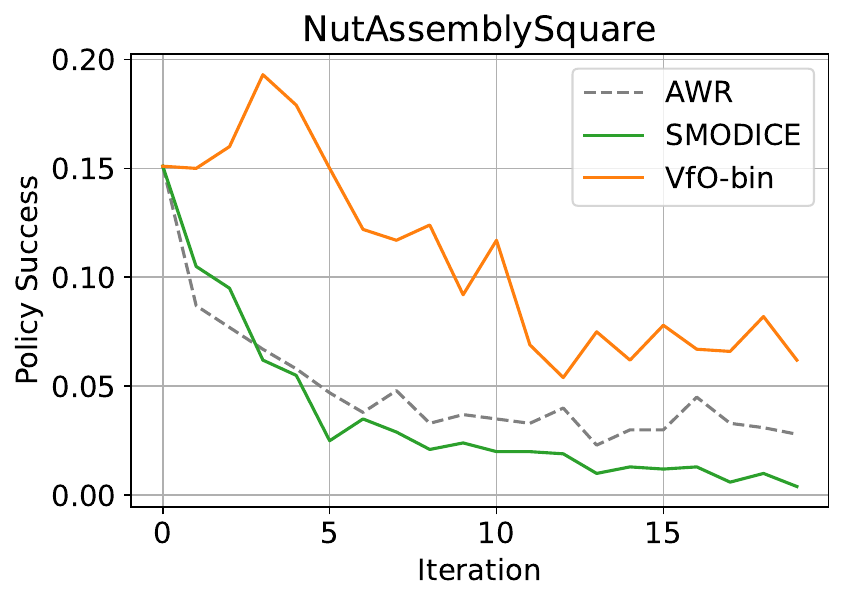}
    \caption{Self-improvement experiments for Robomimic tasks. Starting from different initial performances we can observe whether performance increases or not. Results are mixed, VfO-bin is good on two out of three tasks. Given AWR's dependency on informative rewards (here: sparse),  VfO-bin may slightly outperform offline RL in this setting.}
    \label{fig:robomimic_zth}
\end{figure*}

To confirm the validity of our SIBench proxy, we run self-improvement experiments akin to what we envisioned in the introduction: After learning an initial policy from a seed dataset, we collect 1,000 episodes to form a new dataset for the next iteration (and then repeat this process in an improvement loop). We perform 20 iterations in total and seed with data with bad but non-zero performance to avoid regions with low signal-to-noise ratios.
In order to observe correlation with SIBench we pick VfO-bin as our method to benchmark (for which we expect good performance) and use SMODICE as a baseline. We again run AWR (assuming rewards on all data) to compare to a form of oracle performance. Further baselines are provided for D4RL tasks.

\Cref{fig:d4rl_zth,fig:robomimic_zth} show policy performance against self-improvement iteration step.
We can confirm that whenever we see performance improvement in SIBench we also attain self-improvement when collecting data online, confirming the representative power of our offline proxy. For Ant, HalfCheetah, and Walker2D the self-improvement has not converged within the alotted iterations. Interestingly, for Hopper VfO-bin converges roughly where the zero-crossing of the performance lies in SIBench between 1'500 and 2'000. Additionally, self-improvement is also obtained for Robomimic when starting from data that shows positive SIBench improvement.


Overal, VfO-bin clearly outperforms all baselines in this setting and, remarkably, obtains performance similar to the AWR oracle. We want to highlight that this is a highly non-trivial result, bootstrapping imitation learning from observations to mastery via self-collection starting from low quality data is an open problem and to the best of our knowledge has not been demonstrated before.

\section{Limitations and Future Opportunities}

While a key application of IfO targets transfer from direct human provided (third person) demonstrations of a task rather than trained first person control, all presented experiments are limited to consistent embodiment between demonstrations and additional data source. Intermediate steps in this direction might utilize state estimation techniques to map correspondences between robot and human states \citep{luo2024rlif}, but the final goal should remain to exploit existing semantic understanding in pre-trained vision-language and other foundation models \citep{stone2023open,yuan2024robopoint,pmlr-v229-zitkovich23a, Wulfmeier2023FoundationsFT,majumdar2023we}.
The considerable computational requirements of such models renders the iterative offline learning setting we describe in \Cref{sec:benchmarks} more tractable than the pure online learning setting. Furthermore, while we have highlighted shortcomings of current evaluation regimes and sketched a path to scaling IfO, we have not yet done any large scale experiments yet. We intend to address this in future work, incrementally tackling multi-task and multi-domain settings.

\section{Conclusions}
Imitation learning from observation has the potential to become a principal component of large-scale behavior learning. We advance this paradigm by suggesting the use of IfO in conjunction with self-improvement. We provide a novel offline benchmark, which we find to be much more representative of this self-improvement setting when compared to existing benchmarks.
We also present a simple algorithm (VfO) that builds on ideas from SQIL, ORIL, and AWR to effectively train agents by relying on offline RL as the mechanism to learn to imitate. Remarkably, across nearly all experiments in our analysis, VfO is competitive with RL from ground-truth rewards when using just a few action-free trajectories to define desired behavior. VfO, and IfO in general, provides an efficient path to real-world behavior learning, where reward function design or annotated demonstrations often becomes the restricting factor.



\section*{Impact Statement}

This paper presents work focused on advancing the fundamental understanding of imitation learning.  Due to its importance, advancements in this areas have the potential to broadly impact various applications of machine learning, including robotics, natural language processing and computer vision.  Improved theoretical foundations can lead to more robust, efficient, and reliable machine learning models, ultimately benefiting society by enabling more effective solutions in these domains.  As with any technological advancement, it is important to consider potential negative impacts, and future work could explore mitigating strategies.  We believe that the long-term benefits of a stronger theoretical understanding of machine learning outweigh these potential risks.


\bibliography{main}
\bibliographystyle{icml2025}

\newpage
\appendix
\onecolumn

\section{Implementation Details}
\label{app:implementation}
The policies are simple MLPs with two hidden layers of either 512 units for D4RL or 1024 units for Robomimic. For the VfO algorithms we set the temperature parameter $\lambda$ to $1$ on the D4RL tasks and to $0.1$ on the Robomimic tasks (see \Cref{app:hyperparameters} for an overview of parameters and a sensitivity analysis).
For the methods that learn value functions, we use a target network that gets updated every 200 steps.
Some additional implementation details are worth pointing out:
We use a multi-scale encoder similar to the one used in the Perceiver Actor-Critic model \citep{springenberg2024offline} that circumvents issues with saturation of non-linearities or insensitivity to lower amplitude signals. This setting reduces dependence on exact input normalisation and can simplify later extensions to multi-task scenarios.
Instead of continuous action predictions, we discretize the actions in 101 uniformly spaced bins for which we learn a categorical distribution as is common for recent transformer architectures in control domains \citep{reed2022generalist}. 
However, we apply a Gaussian kernel on the last layer to provide sufficient inductive bias in the low data regime.
For tasks with terminations, we further bootstrap the values by assuming the agent to continue receiving the same reward, i.e. $v(s_t)=\frac{r_t}{1-\gamma}$. This increases the effect of terminating states and improves performance on terminating tasks.
All algorithms are trained for 1e6 learning steps.

\section{Baselines Details}
\label{app:baselines}
\paragraph{BCO} For the BCO \citep{torabi2018behavioral} baseline, we train an inverse dynamics model to predict $p(a_t|s_{t}, s_{t+1})$ on the background data and subsequently use this to label the demonstrations. After this, regular BC learning is done on a $50/50$ mixture of the background data and the now action annotated demonstrations. To implement this, we used the same architecture for the dynamics model as we used for the policies in our experiments.

\paragraph{SMODICE} Like in the paper that proposes SMODICE \citep{pmlr-v162-ma22a}, we first train a discriminator network to distinguish between the expert and background data based on the observations/states. We apply early stopping at 10000 steps as we found this to be beneficial in preliminary experiments. Subsequently, we learn the value function as suggested and derive weights for weighted BC. Unlike the original paper, we don't apply entropy regularization to policy and don't apply gradient penalization to the discriminator. The SMODICE discriminator has the same architecture as the discriminator for VfO-disc.

\paragraph{DILO} This algorithm learns a state-state value function that takes as input \emph{two} adjacent states and does policy improvement via AWR. We implemented the same loss as in \citet{sikchi2024dual}. Unlike in that work, we found the orthogonal gradient method from \citet{mao2024odice} to lead to worse learning stability than simply using the \emph{true-gradient} update in which no target networks or stop-gradients are used. Despite not using the orthogonal gradient update, DILO was still the strongest baseline for the bimodal data. We hypothesize that this difference in results could be due to architectural differences like the discretized actions and multi-scale encoder that we used. For the AWR part of the algorithm we use a temperature of $10$ -- note that this would be a setting of $0.1$ in the notation of \citet{sikchi2024dual} where the parameter $\tau$ is the inverse of our temperature parameter $\lambda$. We found that increasing the temperature further led to more stable results on the Robomimic but at the cost of essentially turning the algorithm into BC. Otherwise, the settings for this baseline are the same as for the other methods.

\section{Extended Related Work}
\label{app:related_extended}

In addition to the areas described in the main paper, our work strongly relates to and directly builds on research on self-improvement including fundamental reinforcement learning research. The classical idea to have a policy generate its own data to learn and adapt has a long-standing history in RL \citep{sutton2018reinforcement}. 
More recently various works explicitly split the data generation and learning processes \citep{riedmiller2022collect_infer} via model-based \citep{Matsushima2020DeploymentEfficientRL} and model-free RL approaches \citep{Lampe2023MasteringSO,bousmalis2024robocat,springenberg2024offline}. While many of these works rely on externally defined reward functions, related signals, or vision-language models as reward sources \citep{Ma2024VisionLM}, ours directly uses demonstration data to define optimal behaviour \citep{Abbeel2004ApprenticeshipLV}.
Self-improvement research has further gained strong relevance for other foundation model applications such as language modelling \citep{Huang2022LargeLM,Choi2024SelfImprovingRP}.

\section{Hyperparameters}
\label{app:hyperparameters}
\begin{figure}[ht]
    \centering
    \includegraphics[width=.4\linewidth]{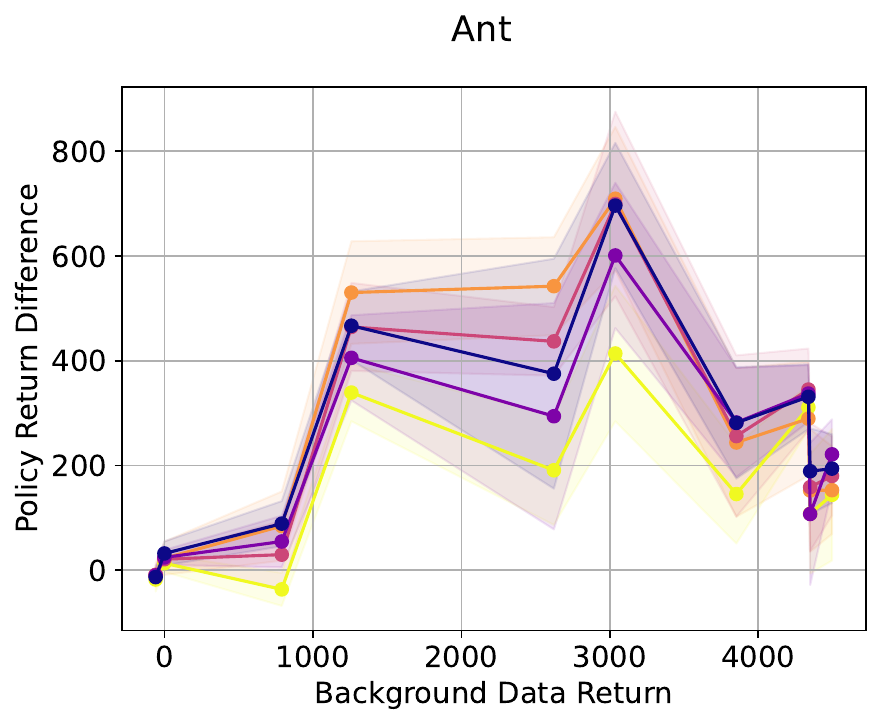}
    \includegraphics[width=.4\linewidth]{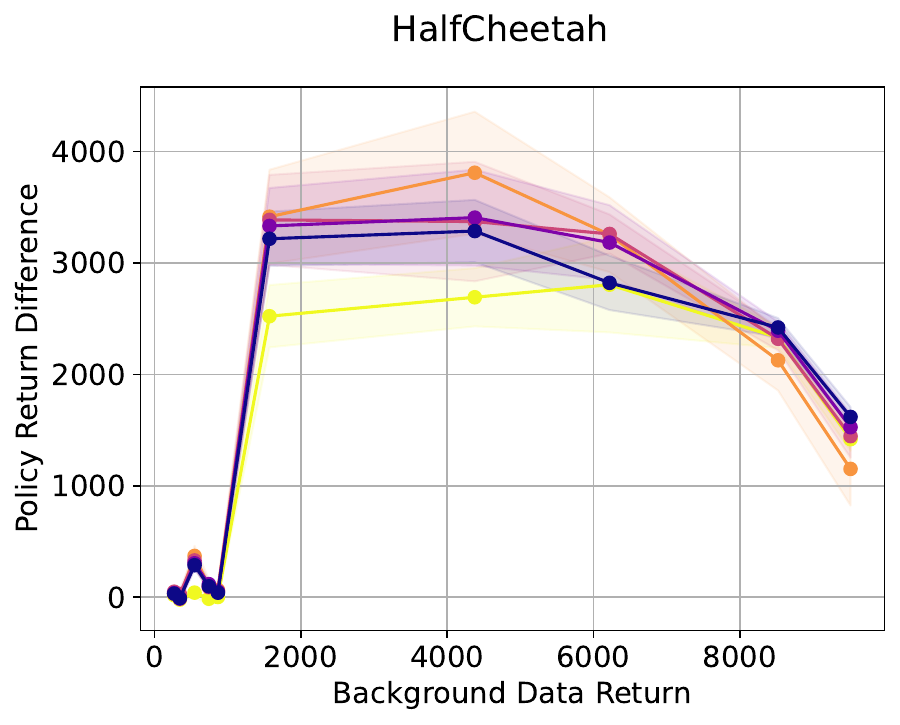}
    \includegraphics[width=.4\linewidth]{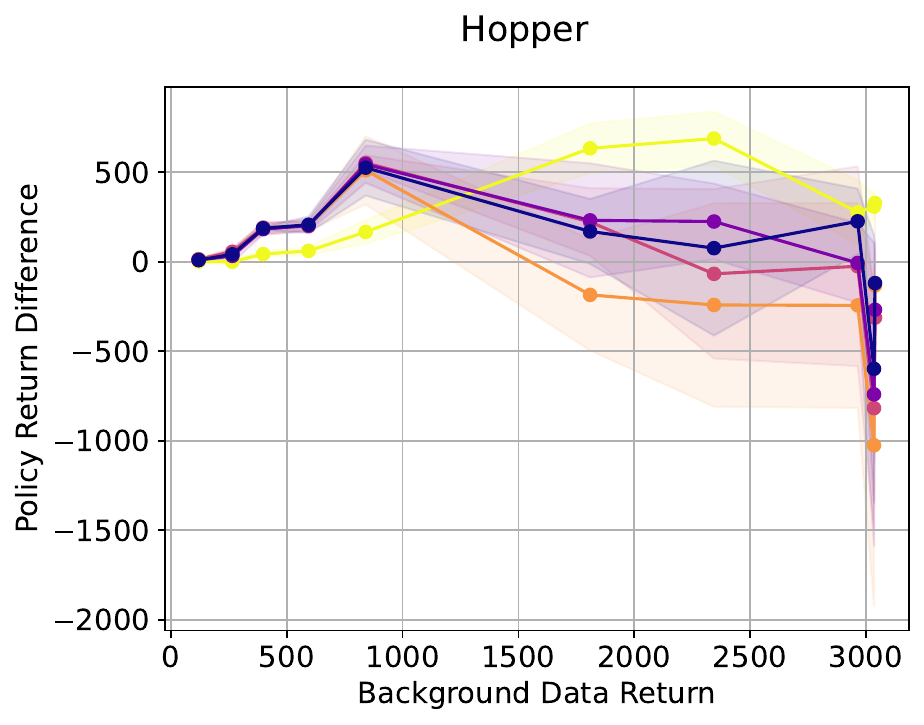}
    \includegraphics[width=.4\linewidth]{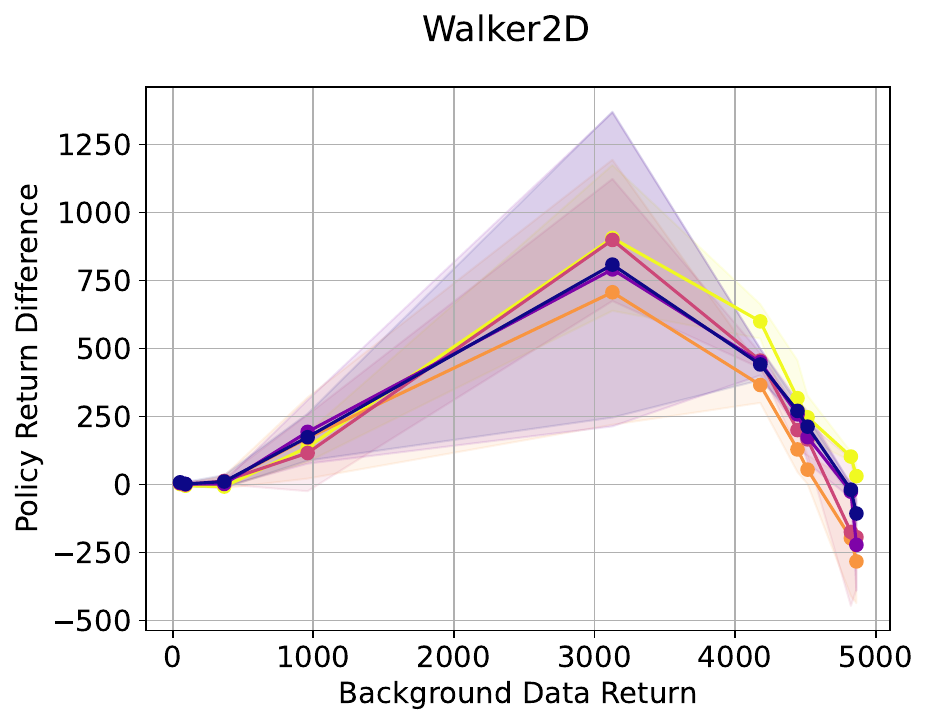}
    \includegraphics[width=.6\linewidth,trim={0 0 0 3cm},clip]{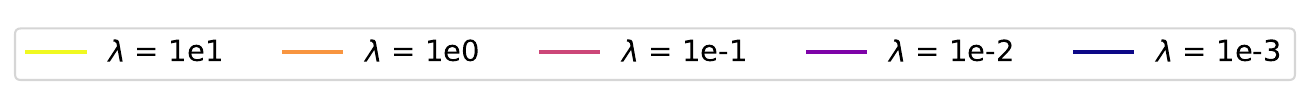}
    \caption{Difference in cumulative return of VfO-bin on D4RL tasks using the SIBench data for different temperature parameters and mixing parameter $0.5$. We plot the average return of the trained policy against the return in the background data. We can observe a fairly wide range of hyperparameter settings leading to improvement.}
    \label{fig:d4rl_irlb_temp}
\end{figure}

\begin{figure}[ht]
    \centering
    \includegraphics[width=.4\linewidth]{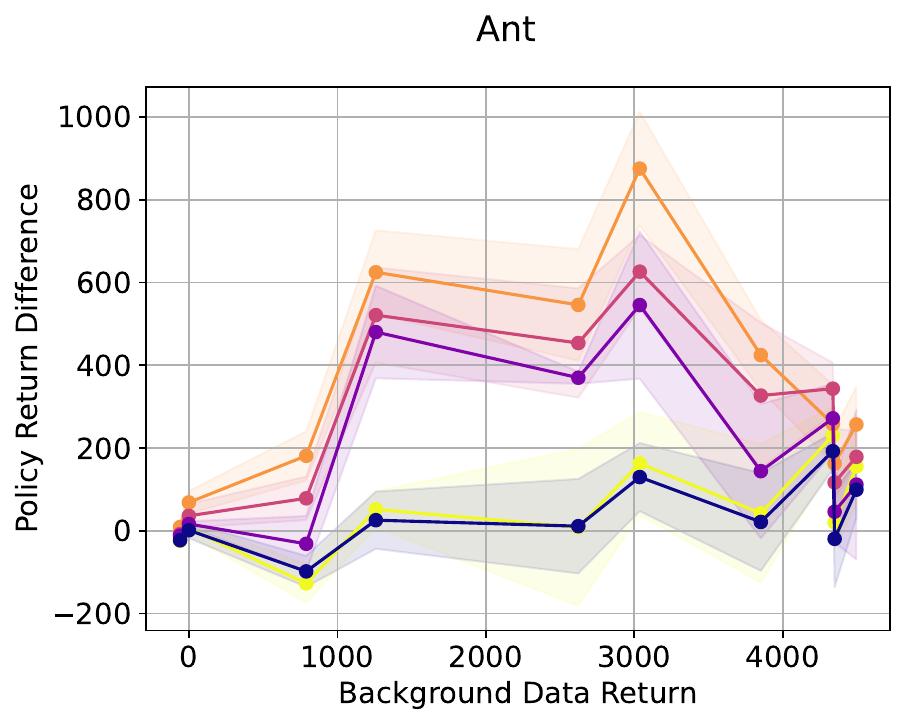}
    \includegraphics[width=.4\linewidth]{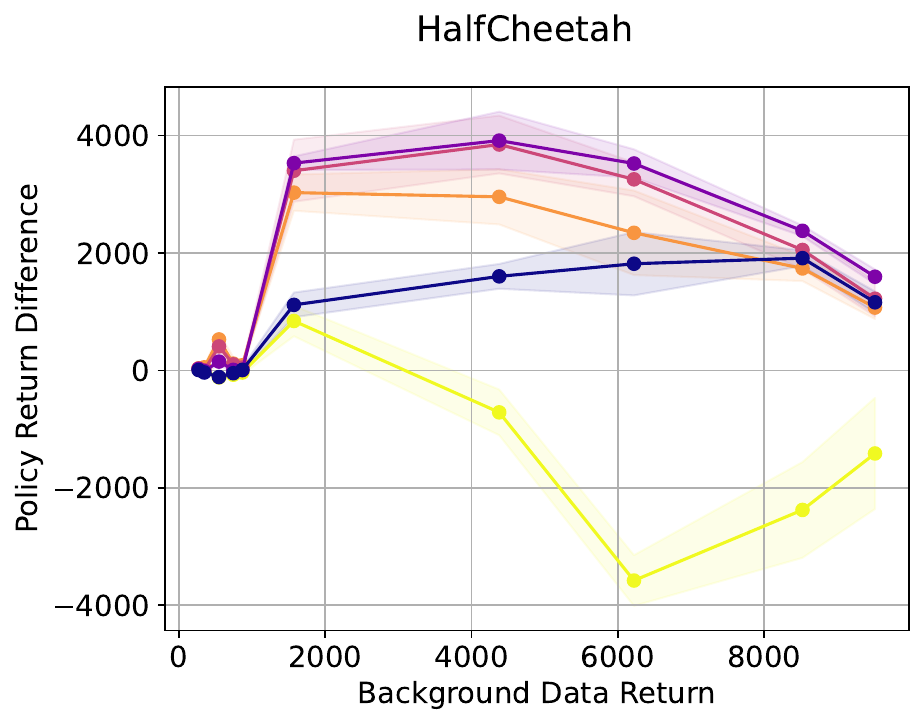}
    \includegraphics[width=.4\linewidth]{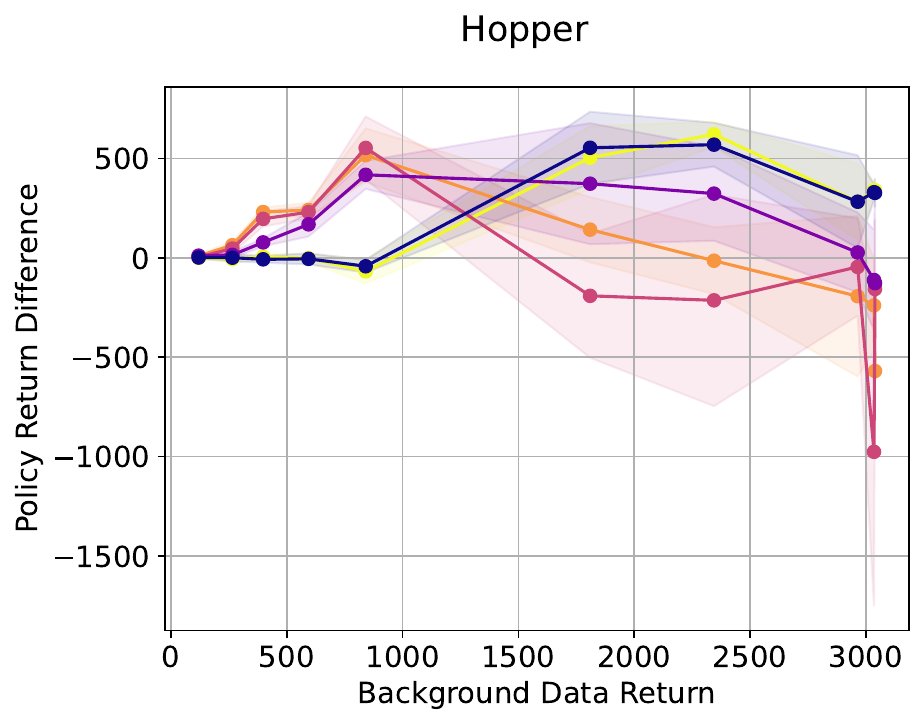}
    \includegraphics[width=.4\linewidth]{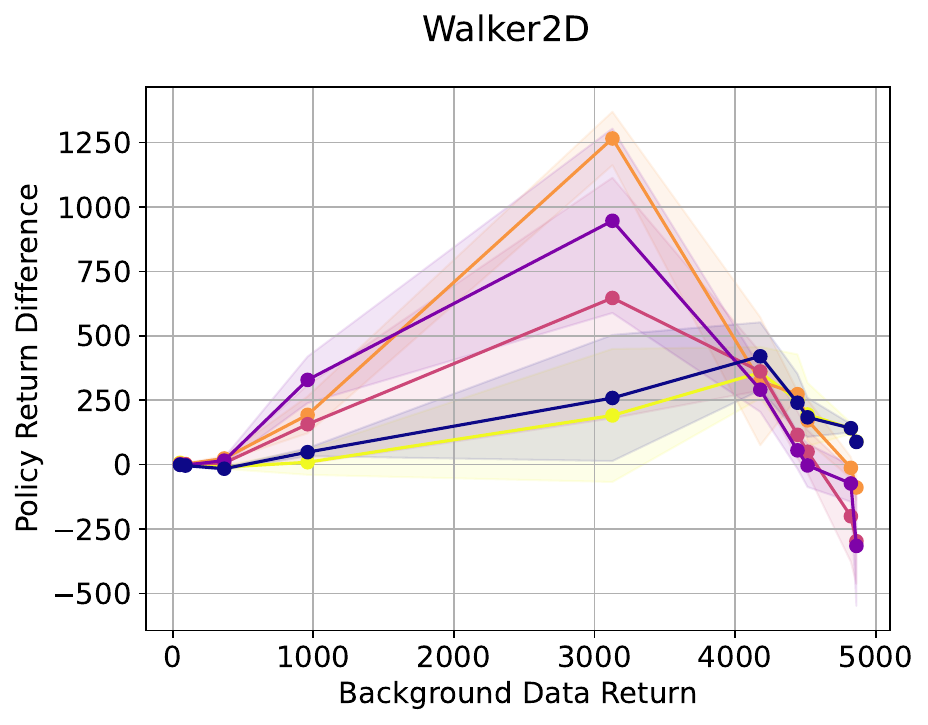}
    \includegraphics[width=.6\linewidth,trim={0 0 0 3cm},clip]{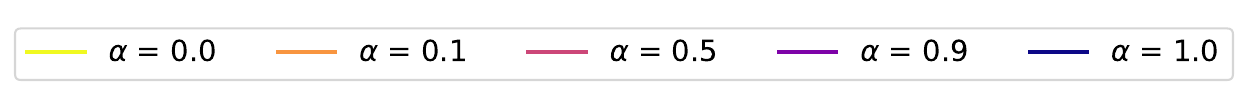}
    \caption{Difference in cumulative return of VfO-bin on D4RL tasks using the SIBench data for different mixing parameters and temperature $1.0$. We plot the average return of the trained policy against the return in the background data. Except for Hopper, we can observe that picking a parameter between 0.1 and 0.9 yields consistent improvement.}
    \label{fig:d4rl_irlb_mix}
\end{figure}

\Cref{tab:hyper} provides an overview of the employed parameters. Further \Cref{fig:d4rl_irlb_temp,fig:d4rl_irlb_mix} provide hyperparameter ablation for the temperature $\lambda$ and the mixing parameter $\alpha$. They confirm that for both there is a wide range of parameters that enable improvement.

\begin{table}[h]
    \centering
    \begin{tabular}{l c}\toprule
     \textbf{Hyperparameter}    & \textbf{Value} \\
     \midrule
      learning rate   &  3e-4 \\
      batch size      & 256 \\
      MLP layers D4RL & (512, 512)\\
      MLP layers Robomimic & (1024, 1024)\\
      target network update period & 200\\
      weight decay D4RL & 0\\
      weight decay Robomimic & 0.1\\
      $\lambda$ (temperature) D4RL & 1.0 \\
      $\lambda$ (temperature) Robomimic & 0.1 \\
      $\alpha$ (mixture ratio) & 0.5\\
      $\gamma$ (discount ratio)  & 0.99\\
      \bottomrule
    \end{tabular}
    \caption{Shared Algorithmic Hyperparameters}
    \label{tab:hyper}
\end{table}

\section{Absolute Return Plots}
\label{app:abolute}
\begin{figure}[ht]
    \centering
    \includegraphics[width=.4\linewidth]{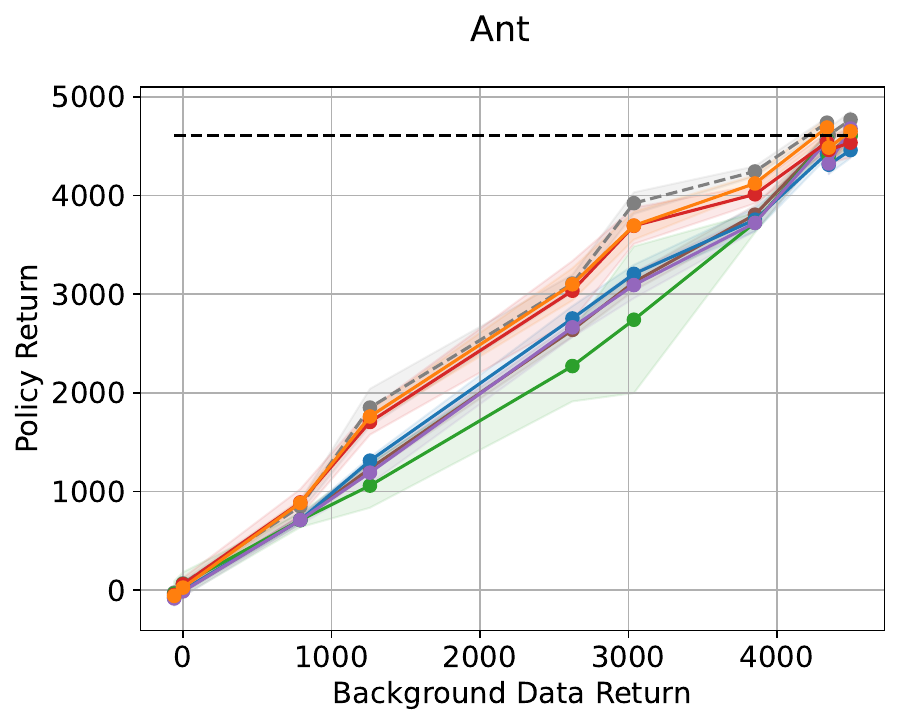}
    \includegraphics[width=.4\linewidth]{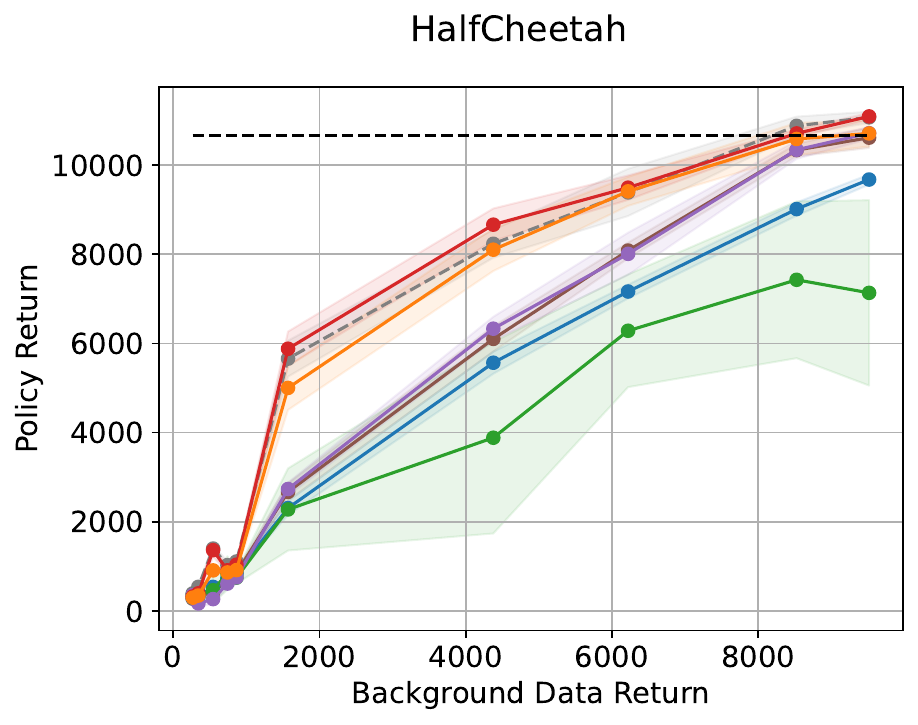}
    \includegraphics[width=.4\linewidth]{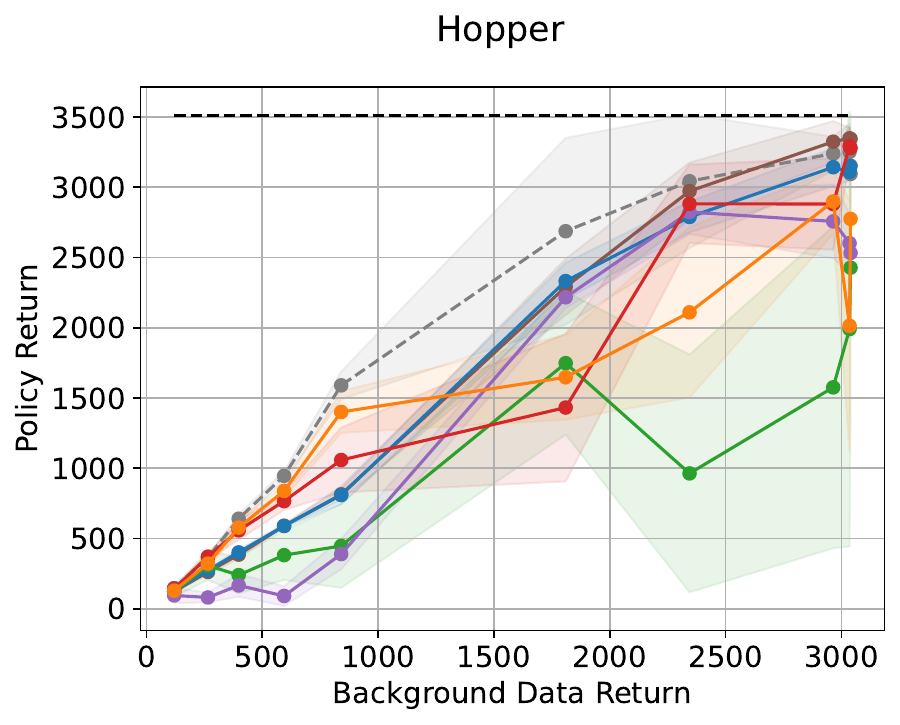}
    \includegraphics[width=.4\linewidth]{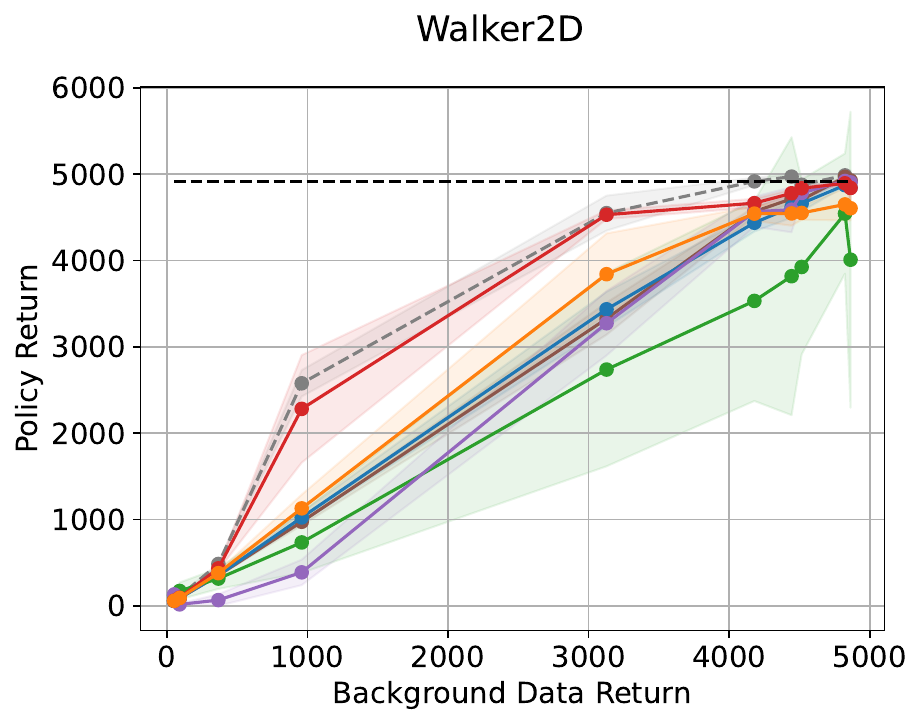}
    \includegraphics[width=.7\linewidth,trim={0 0 0 3cm},clip]{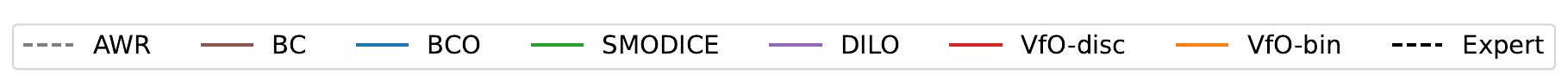}
    \caption{Cumulative return of various algorithms on D4RL tasks using the SIBench data. We plot the average return of the trained policy against the return in the background data.
    AWR, VfO-disc, VfO-bin all show good improvement across the spectrum of \background{} data with the oracle AWR performing best. The corresponding relative plots can be seen in \Cref{fig:d4rl_irlb}.}
    \label{fig:d4rl_irlb_abs}
\end{figure}

\begin{figure}[ht]
    \centering
    \includegraphics[width=.32\linewidth]{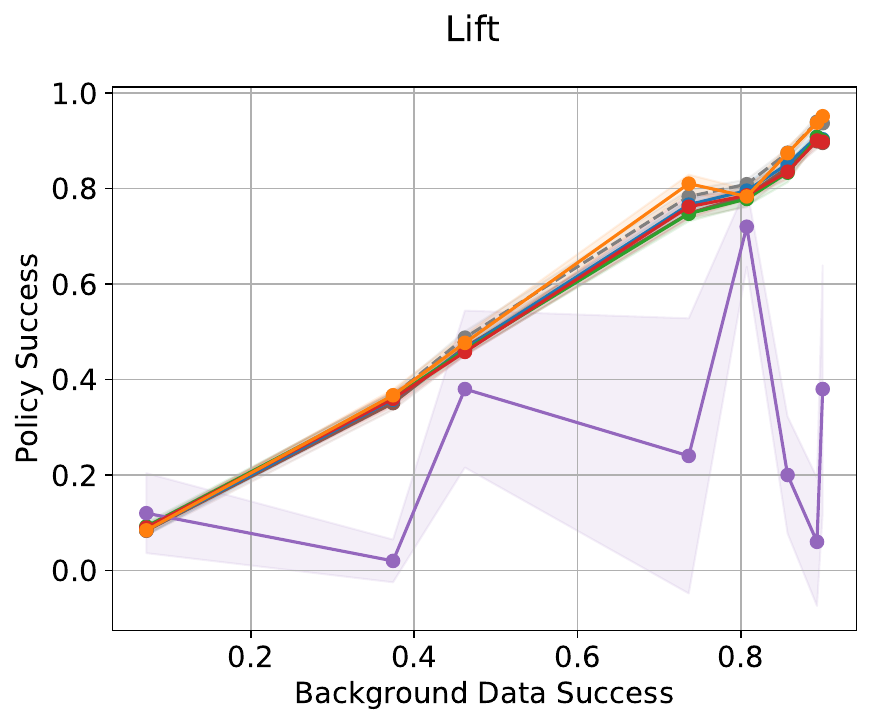}
    \includegraphics[width=.32\linewidth]{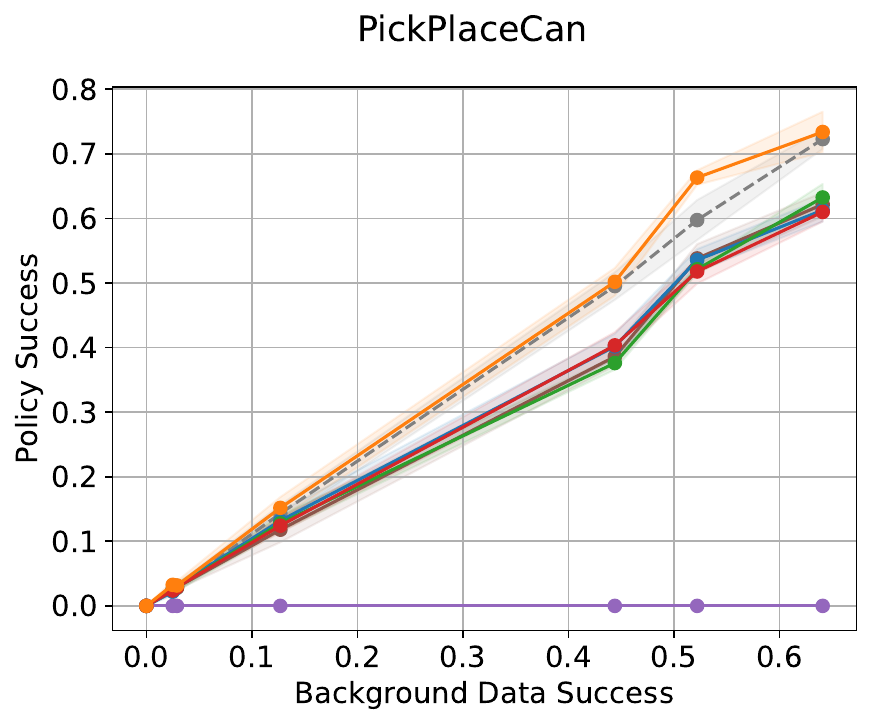}
    \includegraphics[width=.32\linewidth]{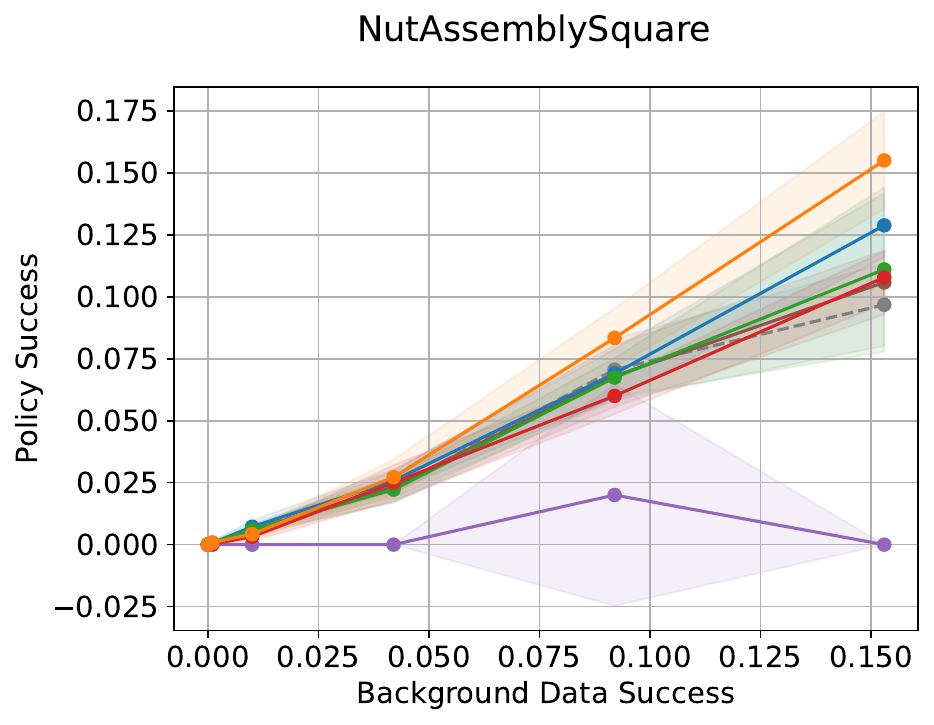}
    \includegraphics[width=.6\linewidth,trim={0 0 0 3cm},clip]{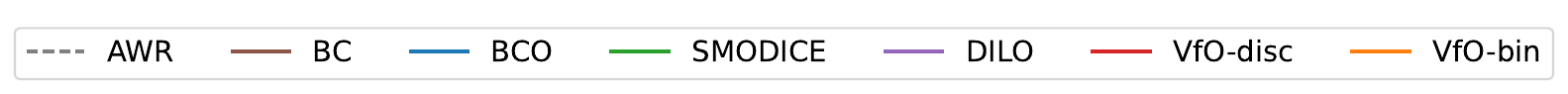}
    \caption{Success rates of various algorithms on Robomimic tasks using the SIBench data. As the absolute performance range is considerably larger than the differences due to highly different initial data quality, relative rankings require a closer look. AWR and VfO-bin mostly yield good improvement. The corresponding relative plots with better resolution can be seen in \Cref{fig:robomimic_irlb}.}
    \label{fig:robomimic_irlb_abs}
\end{figure}

\begin{figure}[ht]
    \centering
    \includegraphics[width=.4\linewidth]{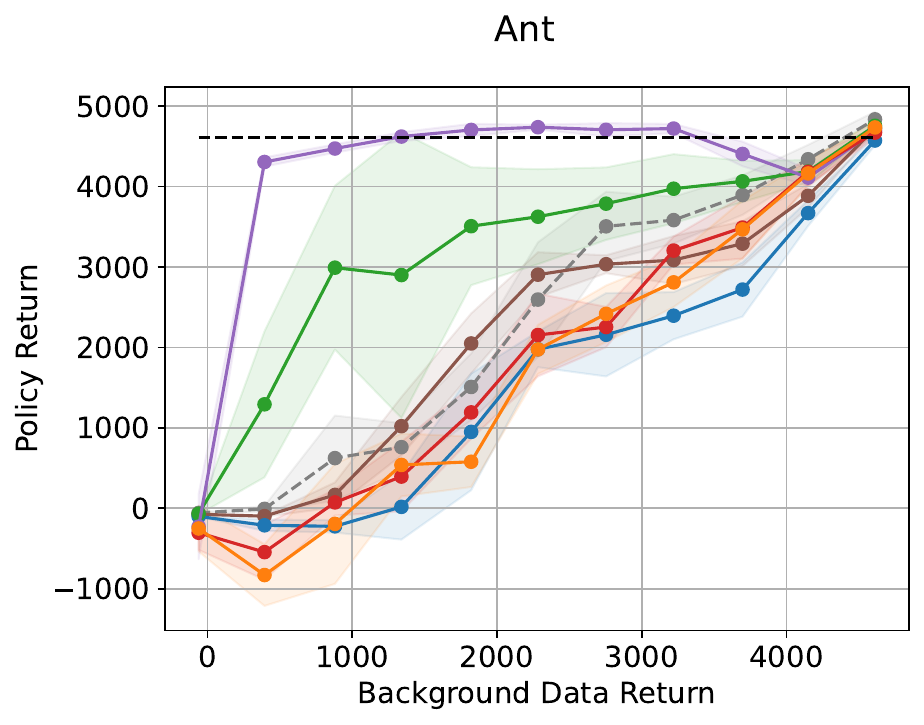}
    \includegraphics[width=.4\linewidth]{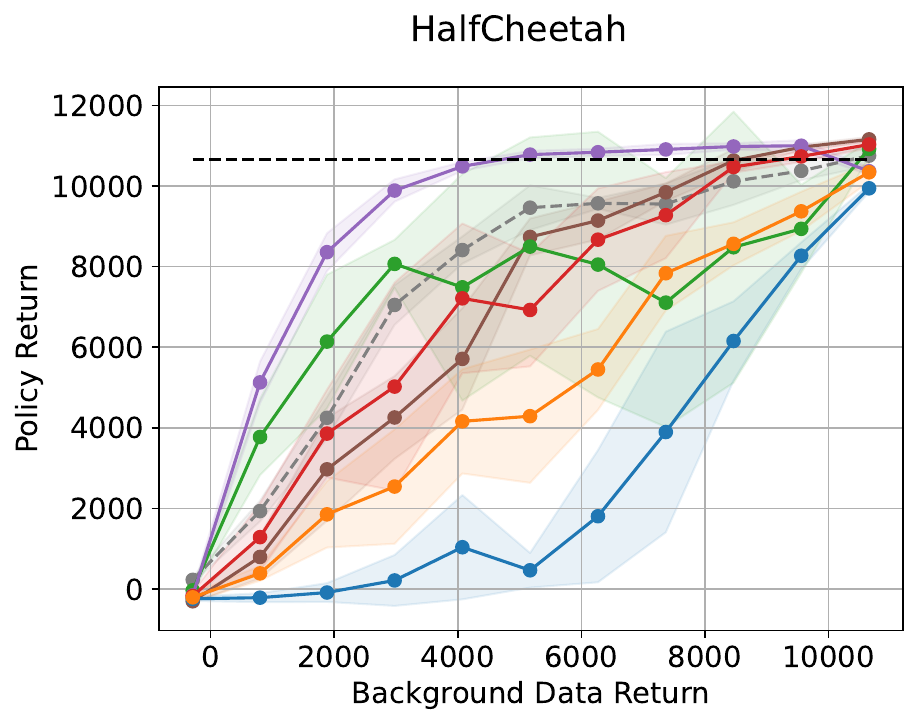}
    \includegraphics[width=.4\linewidth]{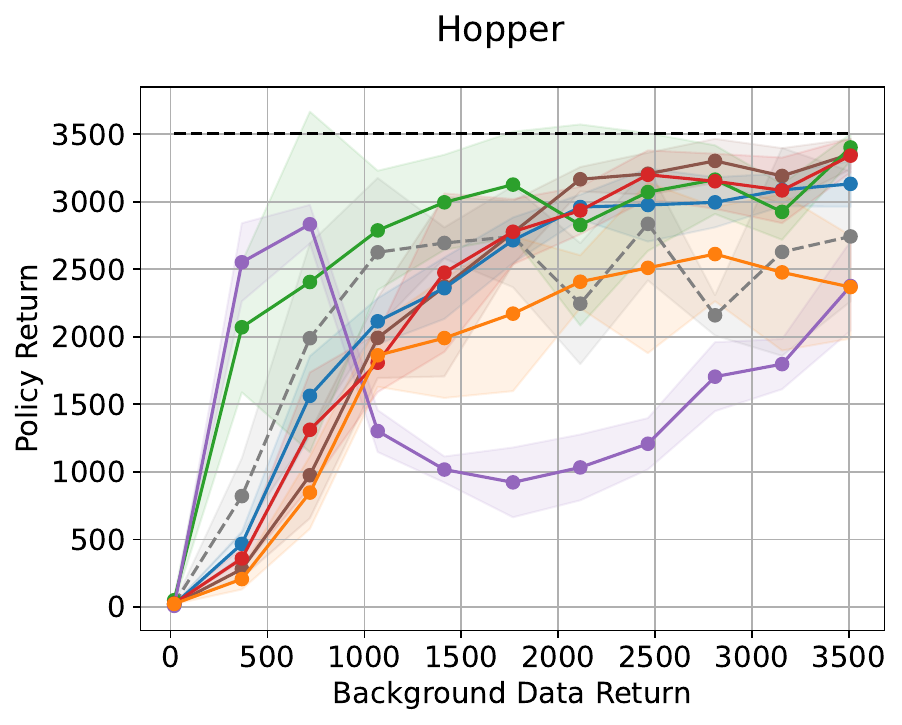}
    \includegraphics[width=.4\linewidth]{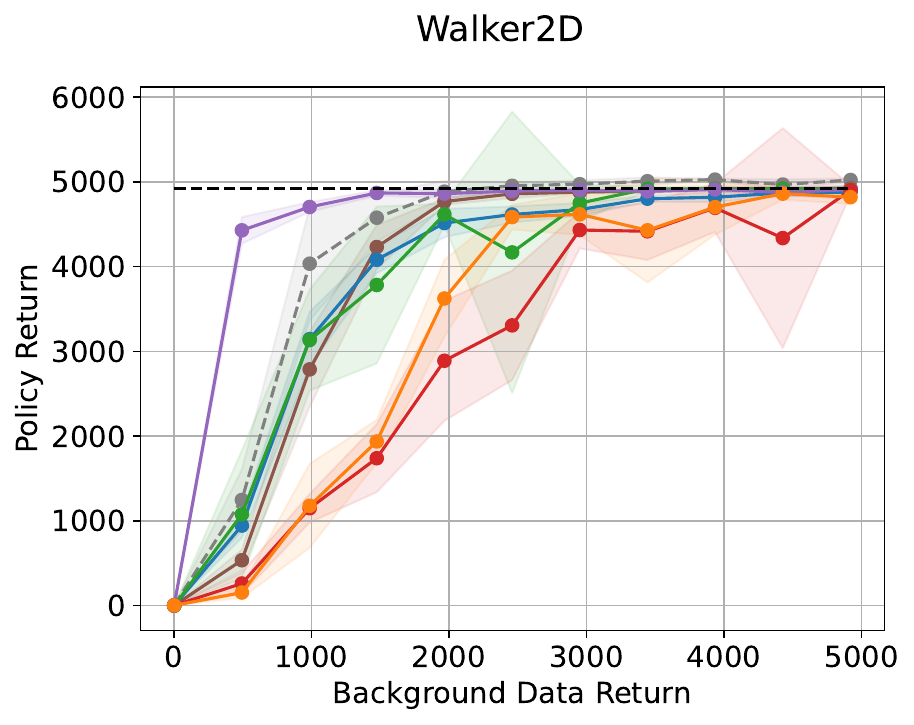}
    \includegraphics[width=.7\linewidth,trim={0 0 0 3cm},clip]{figures/legend_abs.pdf}
    \caption{Cumulative return of various algorithms on D4RL tasks using the bimodal data. As reported in previous work, SMODICE and DILO exhibit strong improvement when the data is composed of a little amount of expert demonstrations. The corresponding relative plots can be seen in \Cref{fig:d4rl_bimodal}.}
    \label{fig:d4rl_bimodal_abs}
\end{figure}

\begin{figure}[ht]
    \centering
    \includegraphics[width=.32\linewidth]{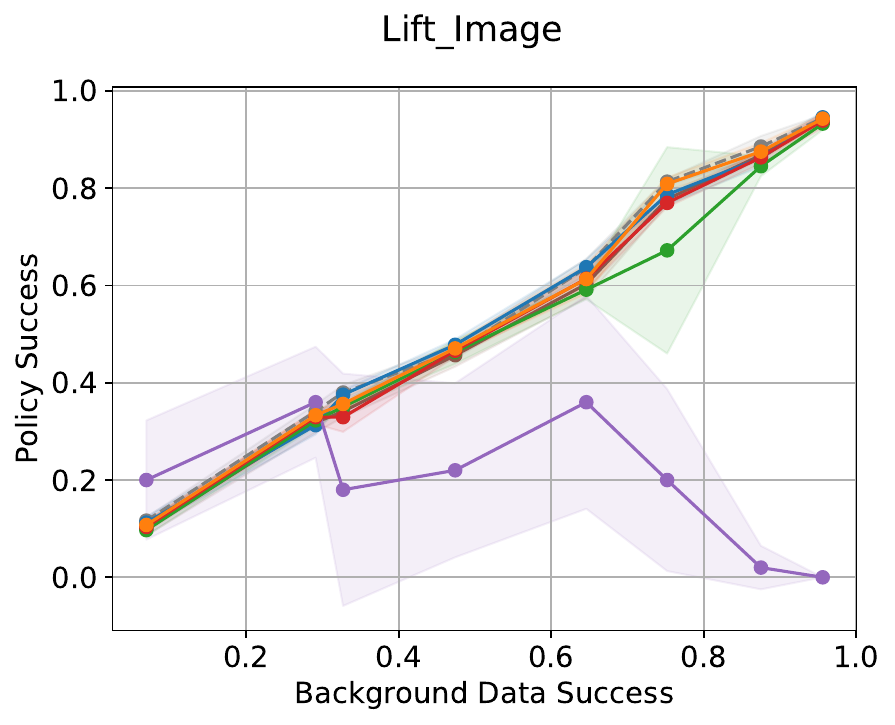}
    \includegraphics[width=.32\linewidth]{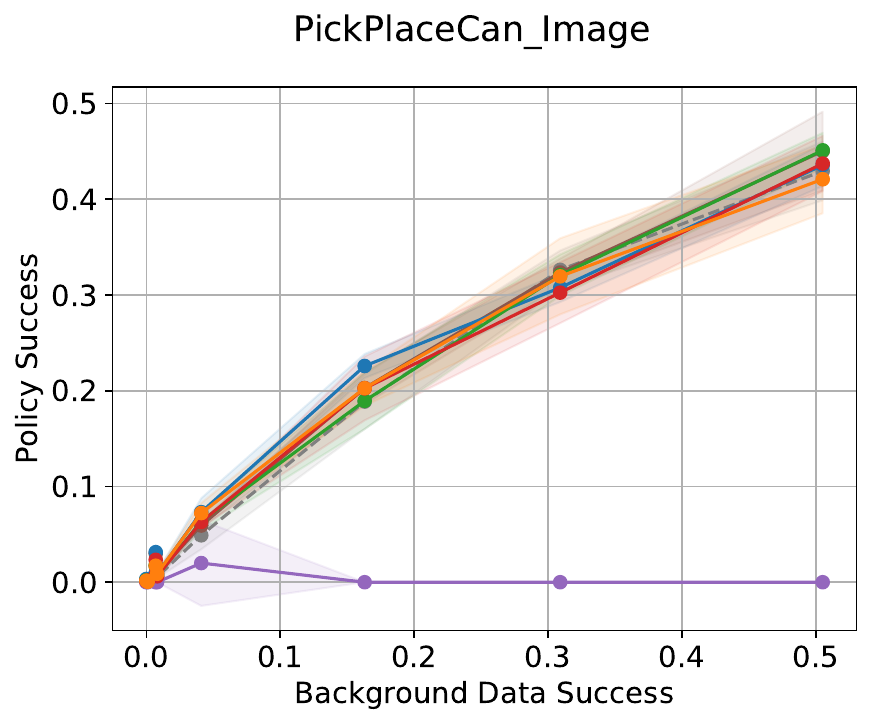}
    \includegraphics[width=.32\linewidth]{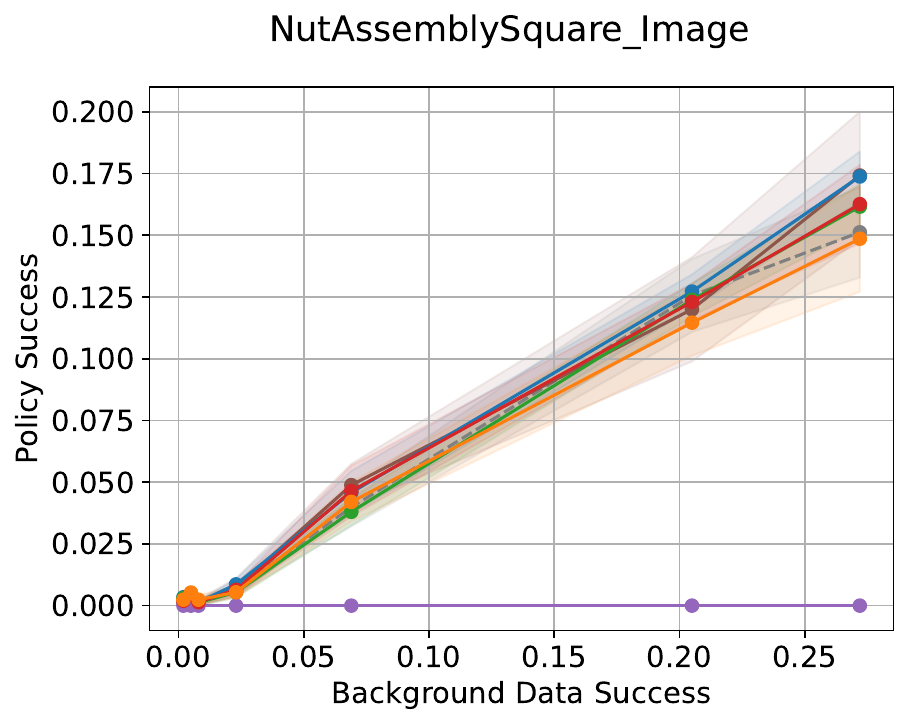}
    \includegraphics[width=.6\linewidth,trim={0 0 0 3cm},clip]{figures/legend.pdf}
    \caption{Success of various algorithms on Robomimic tasks using the SIBench image data. Improvement is difficult to discern in these plots. The corresponding relative plots with better resolution can be seen in \Cref{fig:robomimic_image_irlb}.}
    \label{fig:robomimic_image_irlb_abs}
\end{figure}

The relative plots in \Cref{fig:d4rl_irlb,fig:robomimic_irlb,fig:d4rl_bimodal,fig:robomimic_image_irlb} are not very common in the literature. We thus provide the absolute counterparts in \Cref{fig:d4rl_irlb_abs,fig:robomimic_irlb_abs,fig:d4rl_bimodal_abs,fig:robomimic_image_irlb_abs} even though interpretation might be slightly more difficult.

\section{Offline SQIL with Privileged Expert Actions}
\label{app:asqil}
\begin{figure}[ht]
    \centering
    \includegraphics[width=.4\linewidth]{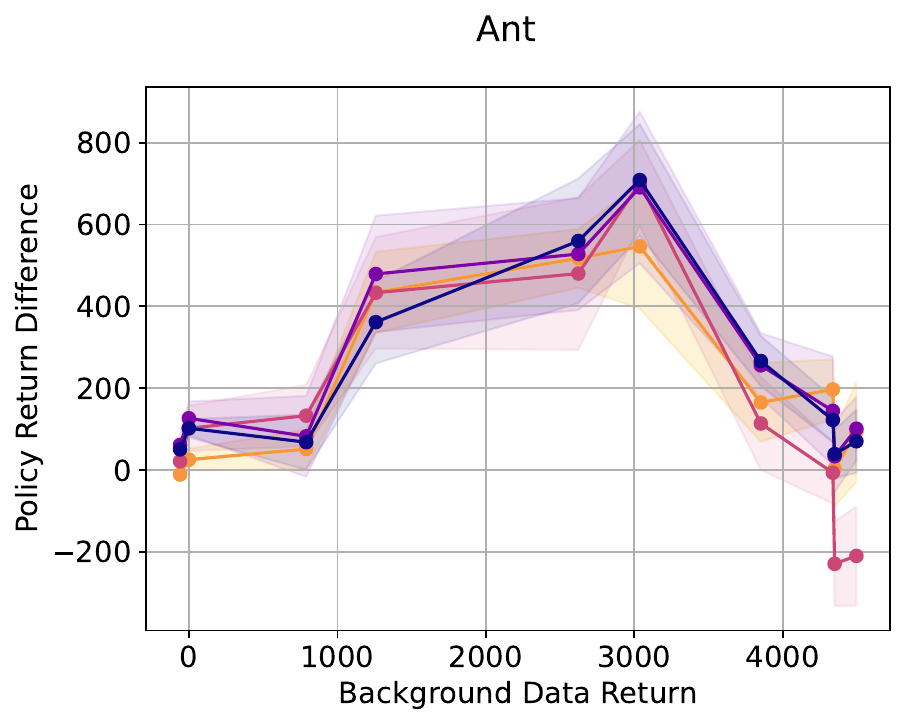}
    \includegraphics[width=.4\linewidth]{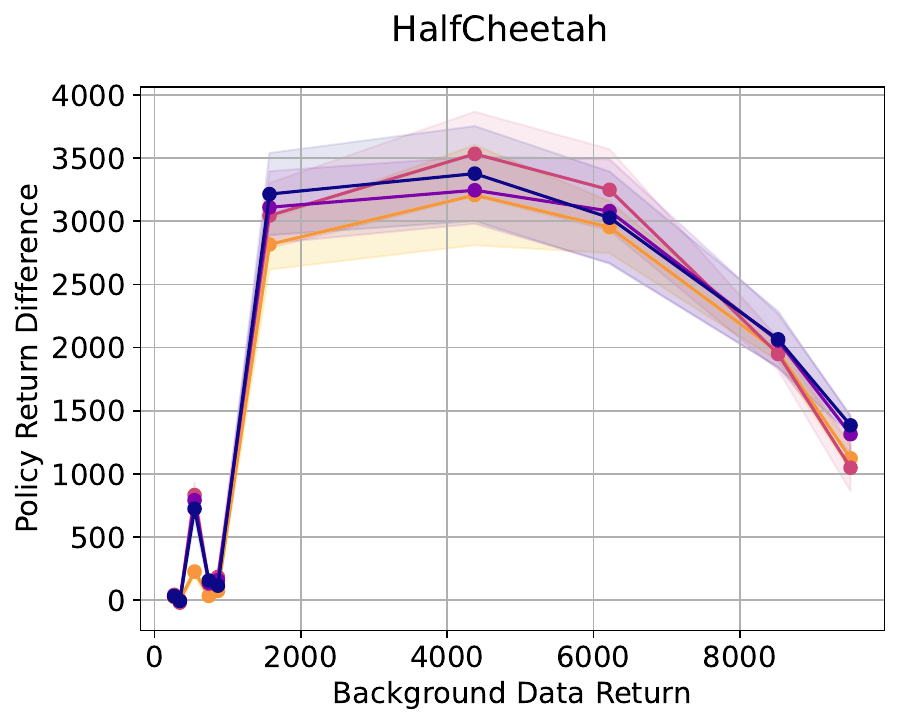}
    \includegraphics[width=.4\linewidth]{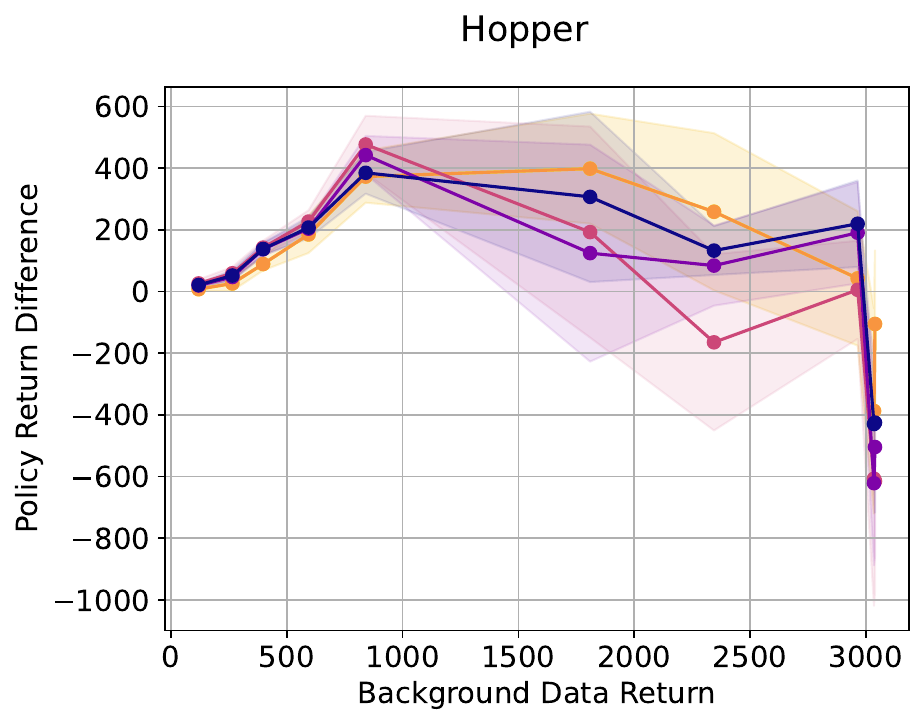}
    \includegraphics[width=.4\linewidth]{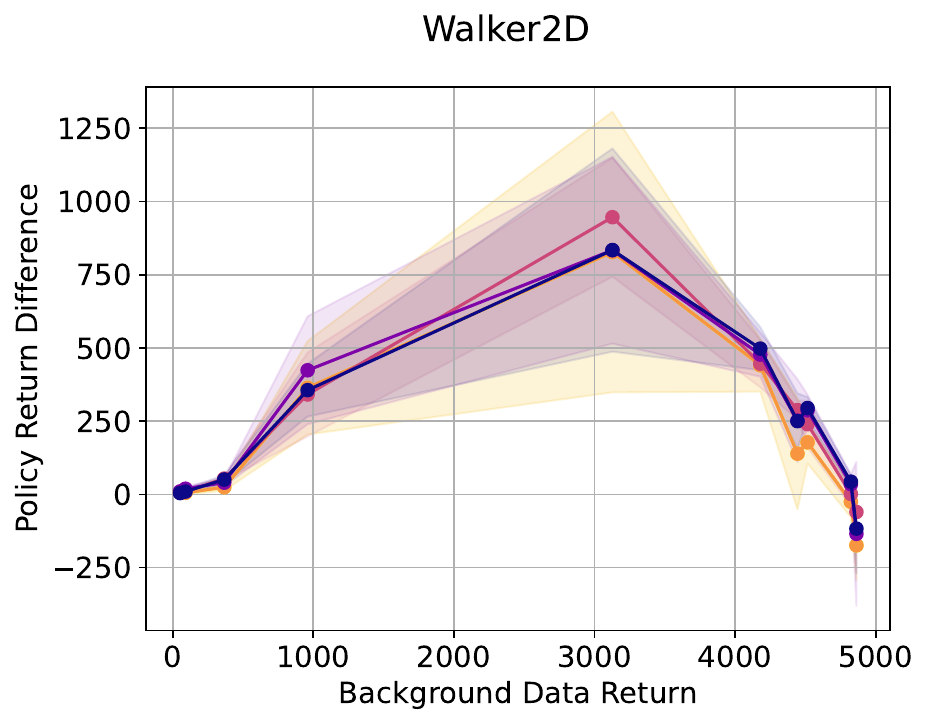}
    \includegraphics[width=.6\linewidth,trim={0 0 0 3cm},clip]{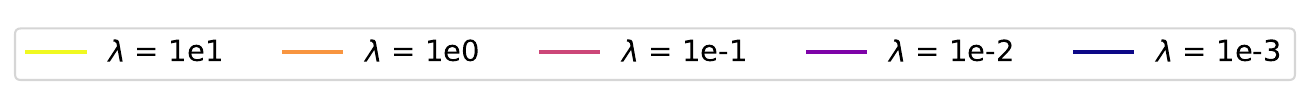}
    \caption{Difference in cumulative return of offline SQIL (with privileged expert actions) on D4RL tasks using the SIBench data for different temperature parameters and mixing parameter $0.9$. We plot the average return of the trained policy against the return in the background data. We can observe a fairly wide range of hyperparameter settings leading to improvement. VfO attains similar or even slightly better improvements (see \Cref{fig:d4rl_irlb_temp}).}
    \label{fig:d4rl_irlb_temp_asqil}
\end{figure}

\begin{figure}[ht]
    \centering
    \includegraphics[width=.4\linewidth]{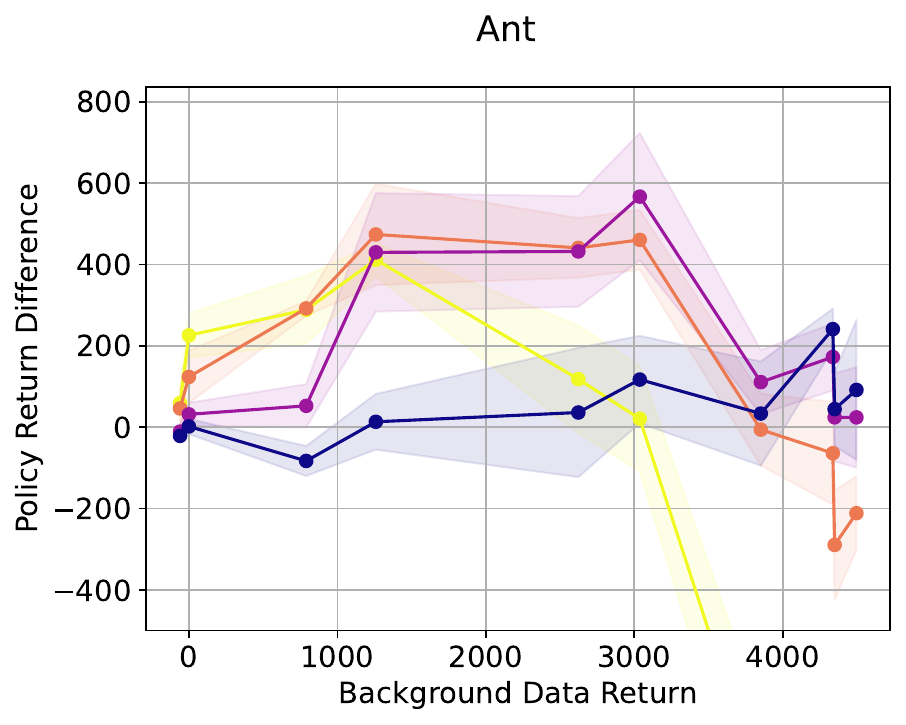}
    \includegraphics[width=.4\linewidth]{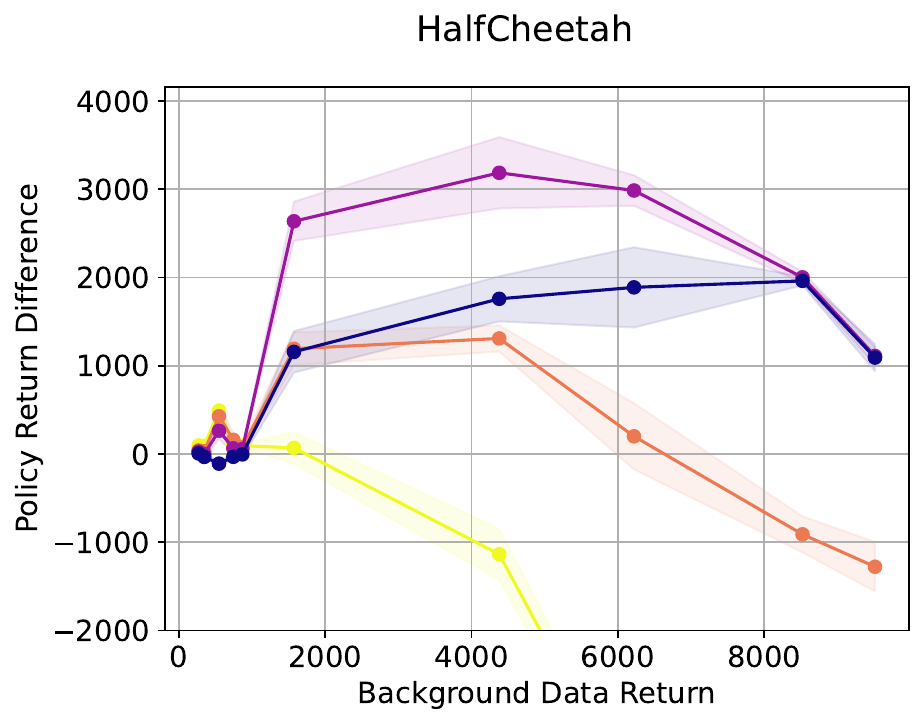}
    \includegraphics[width=.4\linewidth]{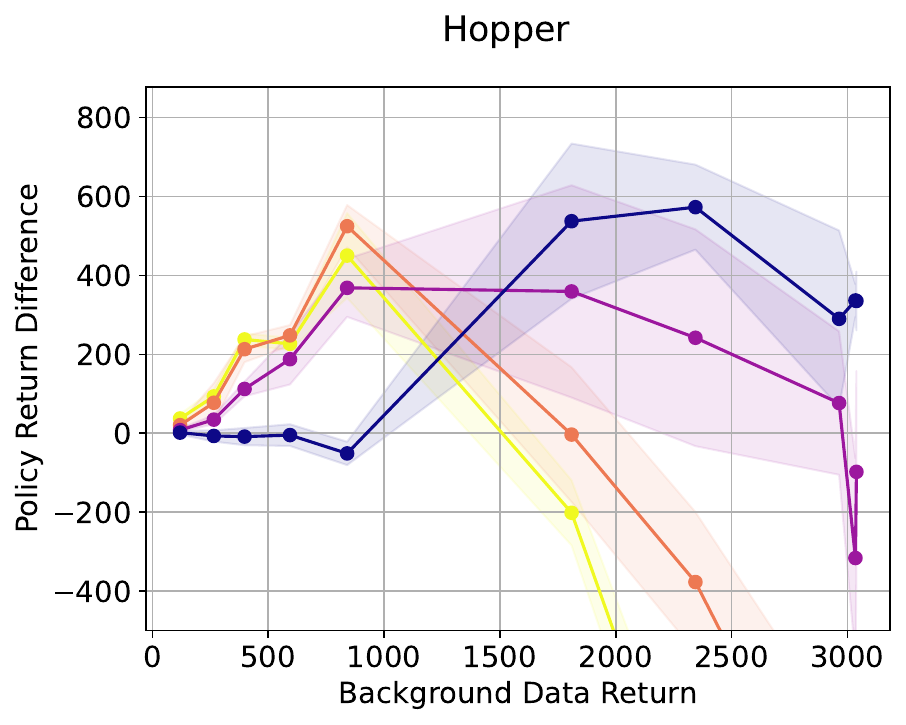}
    \includegraphics[width=.4\linewidth]{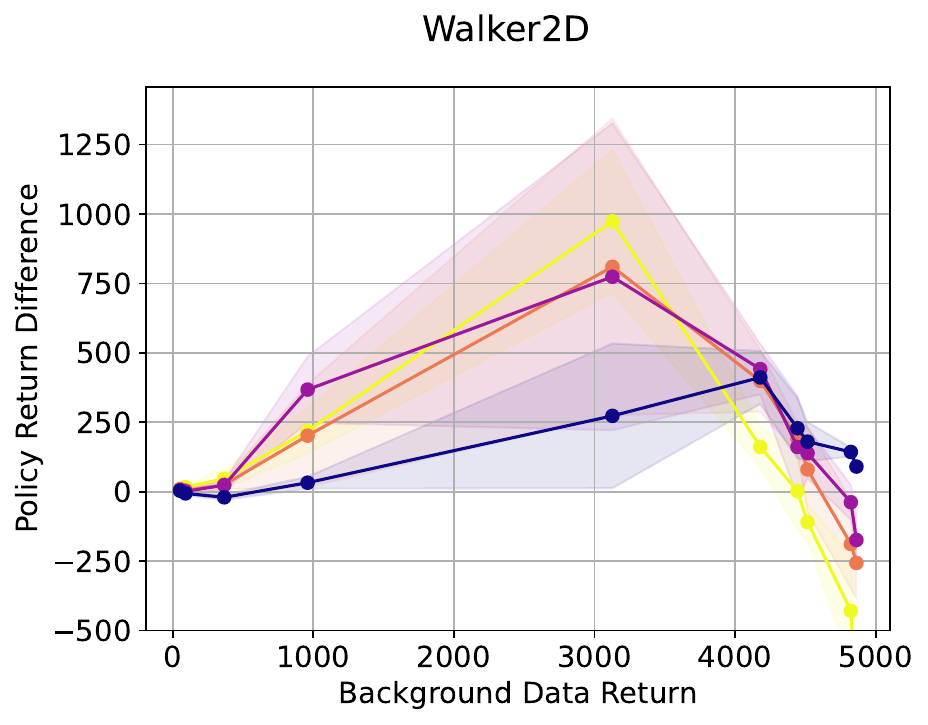}
    \includegraphics[width=.6\linewidth,trim={0 0 0 3cm},clip]{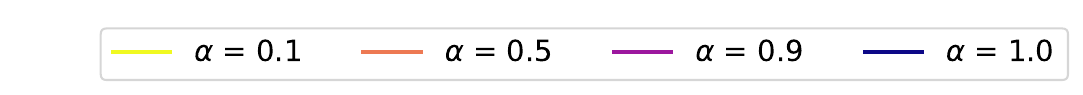}
    \caption{Difference in cumulative return of offline SQIL (with privileged expert actions) on D4RL tasks using the SIBench data for different mixing parameters and temperature $1.0$. We plot the average return of the trained policy against the return in the background data. In comparison to the action-free VfO, we need a higher mixing parameter to avoid overfitting on the scarce expert data.VfO attains similar or even slightly better improvements (see \Cref{fig:d4rl_irlb_mix}).}
    \label{fig:d4rl_irlb_mix_asqil}
\end{figure}

\Cref{fig:d4rl_irlb_temp_asqil,fig:d4rl_irlb_mix_asqil} show results for our offline implementation of SQIL with privileged access to expert actions. We provide hyperparameter ablation for the temperature $\lambda$ and the mixing parameter $\alpha$ and can observe that a higher mixing parameter is required to avoid overfitting on the scarce expert actions. When compared to VfO, we can observe that performance improvement is similar.

\end{document}